\documentclass[runningheads]{llncs}
\usepackage{graphicx}
\usepackage{comment}
\usepackage{amsmath,amssymb} %
\usepackage{color}

\usepackage{hyperref}
\usepackage{url}

\usepackage{booktabs}
\usepackage{xspace}
\newcommand*{\eg}{e.g.\@\xspace}
\newcommand*{\ie}{i.e.\@\xspace}
\newcommand*{\etal}{\textit{et al.}\@\xspace}
\newcommand*{\wrt}{w.r.t.\@\xspace}

\usepackage{float}
\usepackage{braket}
\usepackage{graphicx}
\usepackage{multirow}
\usepackage{adjustbox}
\usepackage{mathtools}
\usepackage{subcaption}
\usepackage{placeins}
\usepackage[ruled,vlined]{algorithm2e}

\DeclarePairedDelimiterX{\norm}[1]{\lVert}{\rVert}{#1}

\newcommand*{\triplet}{\tau     }
\newcommand*{\subj}{\text{subj}}
\newcommand*{\pred}{\text{pred}}
\newcommand*{\obj}{\text{obj}}
\newcommand*{\Loss}{\mathcal{L}}
\newcommand*{\Image}{\mathcal{I}}
\newcommand*{\Graph}{\mathcal{G}}
\newcommand*{\Boxes}{\mathcal{B}}
\newcommand*{\IoU}{\textsc{IoU}}
\newcommand*{\ReLU}{\textsc{ReLU}}
\newcommand*{\Objects}{\mathcal{O}}

\def\tabref#1{table~\ref{#1}}
\def\Tabref#1{Table~\ref{#1}}
\def\appref#1{appendix~\ref{#1}}

\newif\ifappendix
\appendixtrue

\usepackage{amsmath,amsfonts,bm}

\def\figref#1{figure~\ref{#1}}
\def\Figref#1{Figure~\ref{#1}}

\def\secref#1{section~\ref{#1}}

\def\eqref#1{equation~\ref{#1}}
\def\Eqref#1{Equation~\ref{#1}}

\def\1{\bm{1}}

\newcommand{\test}{\mathcal{D_{\mathrm{test}}}}

\def\vb{{\bm{b}}}

\def\ve{{\bm{e}}}

\def\vh{{\bm{h}}}

\def\vn{{\bm{n}}}

\def\vp{{\bm{p}}}

\def\vs{{\bm{s}}}

\def\vu{{\bm{u}}}
\def\vv{{\bm{v}}}

\def\vx{{\bm{x}}}
\def\vy{{\bm{y}}}
\def\vz{{\bm{z}}}

\def\evh{{h}}

\def\evp{{p}}

\def\evs{{s}}

\def\evy{{y}}
\def\evz{{z}}

\def\mW{{\bm{W}}}

\DeclareMathAlphabet{\mathsfit}{\encodingdefault}{\sfdefault}{m}{sl}
\SetMathAlphabet{\mathsfit}{bold}{\encodingdefault}{\sfdefault}{bx}{n}

\begin{document}
\pagestyle{headings}
\mainmatter
\def\ECCVSubNumber{6336}

\title{Explanation-based Weakly-supervised Learning of Visual Relations with Graph Networks}

\titlerunning{Explanation-based Weakly-supervised Learning of Visual Relations}
\author{Federico Baldassarre \and
Kevin Smith \and
Josephine Sullivan \and
Hossein Azizpour}
\authorrunning{F. Baldassarre et al.}
\institute{KTH - Royal Institute of Technology, Stockholm, Sweden\\
\email{\{fedbal,ksmith,sullivan,azizpour\}@kth.se}}
\maketitle

\begin{abstract}
Visual relationship detection is fundamental for holistic image understanding.
However, the localization and classification of (subject, predicate, object) triplets remain challenging tasks, due to the combinatorial explosion of possible relationships, their long-tailed distribution in natural images, and an expensive annotation process.\newline
This paper introduces a novel weakly-supervised method for visual relationship detection that relies on minimal image-level predicate labels.
A~graph neural network is trained to classify predicates in images from a graph representation of detected objects, implicitly encoding an inductive bias for pairwise relations.
We then frame relationship detection as the \textit{explanation} of such a predicate classifier, \ie we obtain a complete relation by recovering the subject and object of a predicted predicate.\newline
We present results comparable to recent fully- and weakly-supervised methods on three diverse and challenging datasets: HICO-DET for human-object interaction, Visual Relationship Detection for generic object-to-object relations, and UnRel for unusual triplets; 
demonstrating robustness to non-comprehensive annotations and good few-shot generalization.
\end{abstract}

\section{Introduction}
\label{sec:intro}

Visual perception systems, built to understand the world through images, are not only required to identify objects, but also their interactions.
Visual relationship detection aims at forming a holistic representation by identifying triplets in the form (subject, predicate, object).
Subject and object are localized and classified instances such as a cat or a boat, and predicates include actions such as \textit{pushing}, spatial relations such as \textit{above}, and comparatives such as \textit{taller than}.

In recent years, we have witnessed unprecedented development in various forms of object recognition; from classification to detection, segmentation, and pose estimation.
Yet, the higher-level visual task of inter-object interaction recognition remains unsolved, mainly due to the combinatorial number of possible interactions \wrt the number of objects. 
This issue not only complicates the inference procedure, but also complicates data collection -- the cost of gathering and annotating data that spans a sufficient set of relationships is enormous. 
In this work, we propose a novel inference procedure that requires minimal labeling thereby making it easier and cheaper to collect data for training. 
\footnote{\footnotesize PyTorch implementation, data and experiments:
\href{https://github.com/baldassarreFe/ws-vrd}{github.com/baldassarreFe/ws-vrd}}\\

Consider the problem of adding a predicate category to a small vocabulary of 20 objects. 
A single predicate could introduce up to $20^2$ new relationship categories, for which samples must be collected and models should be trained.
Moreover, we know that the distribution of naturally-occurring triplets is long-tailed, with combinations such as \textit{person ride dog} rarely appearing~\cite{Peyre_ICCV_2019}. 
This exposes standard training methods to issues arising from extreme class imbalance.
These challenges have prompted modern techniques to take a compositional approach~\cite{Lu_ECCV_2016,Qi_ECCV_2018,Gkioxari_CVPR_2018,Peyre_ICCV_2019} and to incorporate visual and language knowledge~\cite{Lu_ECCV_2016,Plummer_ICCV_2017,Peyre_ICCV_2019}, improving both training and generalization.

\begin{figure}[t!]
    \centering%
    \begin{subfigure}{.245\textwidth}
        \centering%
        \includegraphics[width=\linewidth]{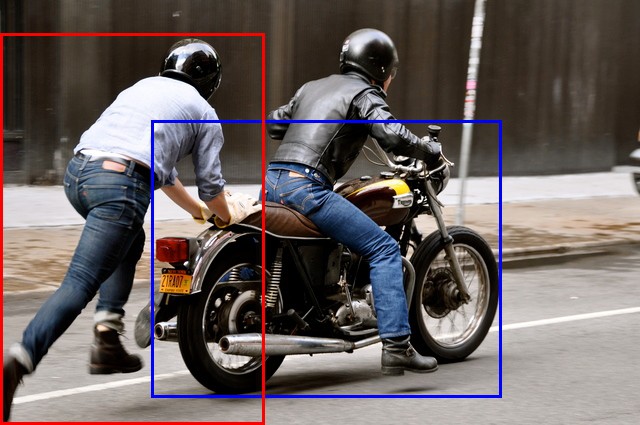}
        \vspace{-.6cm}\caption*{\tiny person push motorcycle}%
    \end{subfigure}\hfill%
    \begin{subfigure}{.245\textwidth}
        \centering%
        \includegraphics[width=\linewidth]{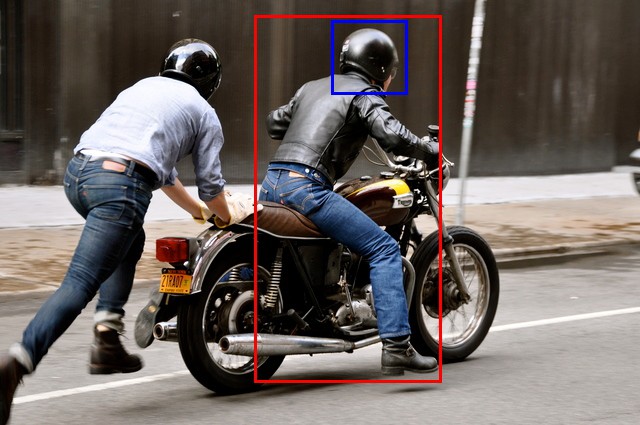}
        \vspace{-.6cm}\caption*{\tiny person wear helmet}%
    \end{subfigure}\hfill%
    \begin{subfigure}{.245\textwidth}
        \centering%
        \includegraphics[width=\linewidth]{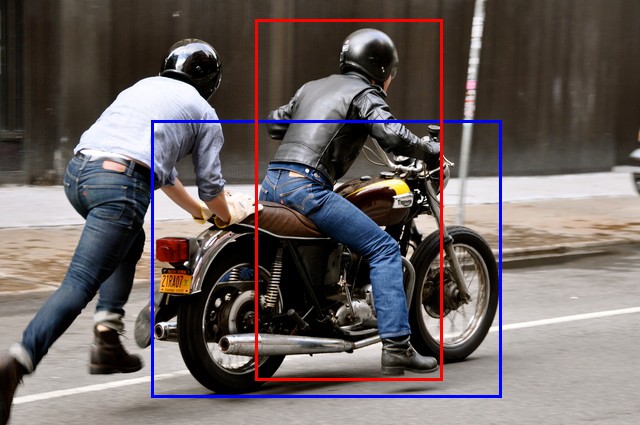}
        \vspace{-.6cm}\caption*{\tiny person drive motorcycle}%
    \end{subfigure}\hfill%
    \begin{subfigure}{.245\textwidth}
        \centering%
        \includegraphics[width=\linewidth]{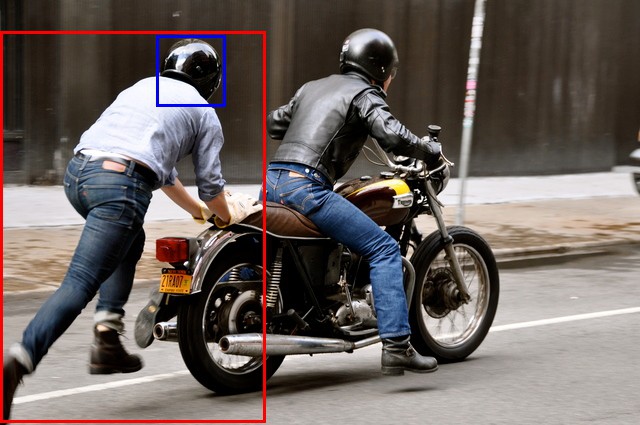}
        \vspace{-.6cm}\caption*{\tiny person wear helmet}%
    \end{subfigure}%
    \caption{\small \textbf{Weakly-supervised relationship detection:} detecting all $\langle\subj,\pred,\obj\rangle$ triplets by training only on weak image-level predicate annotations \textit{\{push, wear, drive\}}}%
    \label{fig:intro-relationship-detection}%
\end{figure}

Although some progress has been made towards recognition of rare triplets, successful methods require training data with exhaustive annotation and localization of $\langle\subj,\pred,\obj\rangle$ triplets. 
This makes weakly-supervised learning a promising research direction to mitigate the costs and errors associated with data collection. 
Nonetheless, we identified only two weakly-supervised works tackling general visual relation detection~\cite{Peyre_ICCV_2017,Zhang_ICCV_2017}, both requiring image-level triplet annotation. 
In this work, we use an even weaker setup for visual relationship detection that relies only on \textit{image-level predicate} annotations (\figref{fig:intro-relationship-detection}).

To achieve that, we decompose a probabilistic description of visual relationship detection into the subtasks of object detection, predicate classification and retrieval of localized relationship triplets. 
Due to considerable progress in object detection, we focus on the last two and use existing pre-trained models for object detection.
For predicate classification, we use graph neural networks operating on a graph of object instances, encoding a strong inductive bias for object-object relations. 
Finally, we use backward explanation techniques to attribute the graph network's predicate predictions to pairs of objects in the input.

\noindent\textbf{Contributions.}
The main contributions of this work are threefold:\\
\noindent I) 
We tackle visual relation detection using a weaker form of label, \ie \textit{image-level predicate annotations} only, 
which reduces data collection cost, is more robust to non-exhaustive annotations, and helps generalization \wrt rare/unseen triplets.\\ 
\noindent II) We propose a novel explanation-based weakly-supervised approach for relationship detection. We believe this is the first work to \textit{(a)} use  weakly-supervised learning beyond object/scene recognition, and  \textit{(b)} employ explanation techniques on graph networks as the \textit{key component} of a relationship detection pipeline.\\
\noindent III) Despite using weaker supervision, we show comparable results to state-of-the-art  methods with stronger labels on several visual relation benchmarks.

\section{Related Works}
\label{sec:related_works}
We are interested in weakly-supervised learning of visual relations. 
We achieve this by employing graph network explanation techniques. 
In this section, we cover the related papers corresponding to the different aspects of our work.

\textbf{Visual Relationship Detection.} 
Visual relation detection involves identifying groups of objects that exhibit semantic relations, in particular (subject, predicate, object) triplets.
Relations are usually either comparative attributes/relative spatial configurations~\cite{Galleguillos_CVPR_2008} which are useful for referral expression~\cite{Mao_CVPR_2016} and visual question answering~\cite{Johnson_CVPR_2017}, or, inter-object interactions~\cite{Sadeghi_CVPR_2011} which is crucial for scene understanding. 
Due to the importance of human-centered image recognition for various applications, many of such works focus on human-object interactions~\cite{Yao_CVPR_2010,Chao_ICCV_2015,Chao_WACV_2018,Qi_ECCV_2018,Gkioxari_CVPR_2018,Zhou_ICCV_2019}. \\
Visual relation detection has been initially tackled by considering the whole relationship triplet as a single-phrase entity~\cite{Sadeghi_CVPR_2011}. 
However, this approach comes with high computational costs and data inefficiency due to the combinatorial space of possible phrases. 
It is therefore important to devise methods that improve data efficiency and better generalize to rare or unseen relations.\\
Most modern works take a compositional approach~\cite{Lu_ECCV_2016,Plummer_ICCV_2017,Peyre_ICCV_2017,Qi_ECCV_2018,Gkioxari_CVPR_2018,Peyre_ICCV_2019}, where objects and predicates are modelled in their own right, which enables better and more efficient generalization. 
Leveraging language through construction of priors, text embeddings, or joint textual-visual embeddings has also been shown to improve generalization~\cite{Lu_ECCV_2016,Plummer_ICCV_2017,Peyre_ICCV_2019}. 
The recent work of Peyre~\etal~\cite{Peyre_ICCV_2019} deals with the combinatorial growth of relation triplets using visual-language analogies.
While this approach generalizes well to unseen combinations of seen entities, it adopts a fully-supervised training procedure that demands a considerable amount of annotated triplets for training.\\
In contrast, our approach improves data efficiency by only requiring image-level predicate labels, and instead learning relation triplets through weakly-supervised learning. 
Our non-reliance on the subject/object entities, in turn, improves generalization to unseen relations as, importantly, we do not require subject/object entities to appear in the training set.

\textbf{Weakly-Supervised Learning.} 
Weakly-supervised learning is generally desirable since it reduces the need for costly annotations. 
It has already proven effective for various visual recognition tasks including object detection \cite{Oquab_CVPR_2015,Bilen_CVPR_2016}, semantic segmentation \cite{Durand_CVPR_2017,Lee_CVPR_2019}, and instance segmentation \cite{Zhou_CVPR_2018,Ge_ICCV_2019}. 
Relationship detection can benefit from weakly-supervised learning even more than object/scene recognition, since the number of possible relation triplets grows quadratically with the number object categories. 
Despite this, weakly-supervised learning of visual relations has received surprisingly less attention than object-centric tasks.

\textbf{Weakly-Supervised Learning of Visual Relations.} 
The early work of Prest~\etal~\cite{Prest_PAMI_2011}, similar to our work, only requires image-level action labels. But Prest~\etal~focused on human-object interactions using part detectors, as opposed to general visual relationship detection. 
More recent works \cite{Peyre_ICCV_2017,Zhang_ICCV_2017} learn visual relations in a weakly-supervised setup where triplets are annotated at the image level and not localized through bounding boxes.
Peyre~\etal~\cite{Peyre_ICCV_2017} represents object pairs by their individual appearance as well as their relative spatial configuration. 
Then, they use discriminative clustering with validity constraints to assign object pairs to image-level labels. 
In~\cite{Zhang_ICCV_2017}, three separate pipelines are used, one for object detection, one for object-object relation classification and the third for object-object pair selection for each relation. 
The softmax output of the latter is then used as an attention mechanism over object pairs to account for the weak labels.
 
Both \cite{Zhang_ICCV_2017,Peyre_ICCV_2017} work with non-localized triplets annotated at the image-level\footnote{It should be noted that \cite{Peyre_ICCV_2017} can be extended to work with only predicate annotations, using a new set of more relaxed constraints.}. 
Our weaker supervision setup, by not requiring subject and object annotations, allows for potentially simpler, more general, and less costly construction of large training datasets using search engines or image captions. 
Furthermore, our method is based on object-centric explanations of graph networks, which sets it apart from previous works on weakly-supervised learning of visual relations.

\textbf{Explanation Techniques.} 
In mission-critical applications such as medical prognosis, a real-world deployment of trained AI systems require explanations of the predictions. 
Thus, many explanation techniques have been developed based on local approximation \cite{Ribeiro_SIGKDD_2016}, game theory \cite{Lundberg_NIPS_2017}, or gradient propagation~\cite{Bach_Plos_2015,Zhou_CVPR_2016,Selvaraju_ICCV_2017}.
Recently, following the success of graph networks, explanation methods have been extended to those models as well \cite{Pope_CVPR_2019,Baldassarre_ICMLW_2019,Ying_NeurIPS_2019}. 
We use graph networks to obtain image-level predicate predictions and then apply graph explanation techniques to obtain the corresponding subject and object in an unsupervised manner.

\textbf{Explanation-based weakly-supervised learning.}
The idea of using explanations to account for weak labels has been previously used for object recognition.
Class Activation Mapping (CAM) uses a specific architecture with fully-convolutional layers and global average pooling to obtain object localization at the average pooling layer~\cite{Zhou_CVPR_2016}. 
\cite{Zhou_CVPR_2018} extends this approach by backpropagating the maximum response of the CAM back to the image space for weakly-supervised instance segmentation.
Grad-CAM \cite{Selvaraju_ICCV_2017} generalizes CAM and extends its applicability to a wider range of architectures by pushing the half-rectified gradient backward and using channel-wise average pooling to obtain location-wise importance. 
Similar to CAM, Grad-CAM is applied to ILSVRC~\cite{russakovsky2015imagenet} for weakly-supervised object localization. 
Finally, \cite{Ge_ICCV_2019} develops a cascaded label propagation setup with conditional random fields and object proposals to obtain object instance segmentation from image-level predictions, using excitation back-propagation~\cite{Zhang_IJCV_2018} for the backward pass.
Our work is an extension to this line of research: we consider a more complicated application, namely visual relationship detection, and use explanation techniques on graph networks.

\section{Method}
\label{sec:method}

Detecting visual relationships in an image consists in identifying triplets $\triplet=\langle \subj, \pred, \obj \rangle$ of subject, predicate and object.
For example, \textit{person~drive~car} or \textit{tree~next~to~building}.
To formalize this, we denote the set of objects in an image by $\Objects$, where each object instance, $i$, has a corresponding bounding box $b_i$ and is categorized as $c_i$ according to a vocabulary of object classes $\{1 \dots C\}$.
Predicates belong to a vocabulary of predicate classes $\{1 \dots K\}$ that include actions such as \textit{eating}, spatial relations such as \textit{next to} and comparative terms such as \textit{taller than}.\\
\noindent With this notation, detecting visual relations from an image $\Image$ corresponds to determining high-density regions of the following joint probability distribution:
\begin{equation}
    P(\triplet|\Image) \triangleq P(c_\subj=c_i,\ k_\pred=k,\ c_\obj=c_j,\ b_\subj=b_i,\ b_\obj=b_j\ | \Image),
\end{equation}
where $c_\subj$ and $c_\obj$ indicate resp. the class of the subject and the object, $k_\pred$ indicates the class of the predicate, $b_\subj$ and $b_\obj$ indicate resp. the location of the subject and the object, and $i, j = 1 \dots |\Boxes|$ index the bounding boxes.\\
\noindent To accommodate weakly-supervised learning, we propose the following approximate factorization based on object detection and predicate classification:
\begin{align}
    &P(\tau | \Image)=\nonumber\\ 
    &\; P(c_\subj=c_i | \Image, b_\subj=b_i) P(c_\obj=c_j | \Image, b_\obj=b_j) &\text{\footnotesize object detection} \label{eq:object-detection}\\
    &\; P(k_\pred=k | \Image) &\text{\footnotesize predicate classification} \label{eq:predicate-classification} \\
    &\; P(b_\subj=b_i, b_\obj=b_j | \Image, k_\pred=k) &\text{\footnotesize likelihood of a pair} \label{eq:visual-relation-explanation} \\
    &\; P(c_\subj=c_i, c_\obj=c_j | k_\pred=k). &\text{\footnotesize prior over relations} \label{eq:prior-over-relations}    
\end{align}

For \eqref{eq:object-detection}, we use an object detection pipeline to localize and classify objects in an image.
The two terms, then, refer to the confidence scores assigned by the object detector to the subject and object of the relationship  (\secref{sec:method-object-detection}).

\Eqref{eq:predicate-classification} corresponds to a predicate classifier that predicts the presence of predicate $k$ in the image.
This component only relies on image-level predicate annotations during training, and does not explicitly attribute its predictions to pairs of input objects.
However, by carefully designing the architecture of the predicate classifier, we introduce a strong inductive bias towards objects and relations, which we can later exploit to recover $\langle\subj,\pred,\obj\rangle$ triplets~(sec.~\ref{sec:method-predicate-classification}).

Given a certain predicate $k$, \eqref{eq:visual-relation-explanation} recovers the likelihood of object pairs to be the semantic subject and object of that predicate.
In other words, we wish to identify \textit{all} possible $(\subj, \obj)$ pairs by their likelihood \eqref{eq:visual-relation-explanation} \wrt a given predicate.
Therefore, we use an explanation technique to compute unnormalized scores that associate predicates to pairs of objects (\secref{sec:method-visual-relation-explanation}).

Term \ref{eq:prior-over-relations}, which we refer to as \textit{prior over relationships}, represents the co-occurrence of certain classes as subjects or objects of a predicate, and the directionality of such relationship. 
For instance, it can indicate that \textit{(person, truck)}, with \textit{person} as the subject, is a more likely pair for \textit{drive} than \textit{(fork, sandwich)}.
As such, this term is optional, and excluding it would be the same as assuming a uniform prior.
However, this term assumes great importance in a weakly-supervised setup, since isolated predicate labels provide no clue on the directionality of the relation between subject and object (\secref{sec:method-prior-over-relations}).

\subsection{Object detection}
\label{sec:method-object-detection}
We use an object detection module to extract a set of objects $\Objects$ from a given image $\Image$. 
We describe each object bounding box by the visual appearance features and the classification scores obtained from the detector. 
These objects will then be used to classify the predicates present in $\Image$ and, later on, 
serve as targets for explanations that identify relevant relationship triplets.
Similar to the weakly-supervised setup of Peyre \etal~\cite{Peyre_ICCV_2017} we assume the availability of pre-trained object detectors~\cite{ren2015faster} as there is substantial progress in that field.

\subsection{Predicate classification}
\label{sec:method-predicate-classification}
\begin{figure}[!b]
    \centering%
    \includegraphics[width=\textwidth]{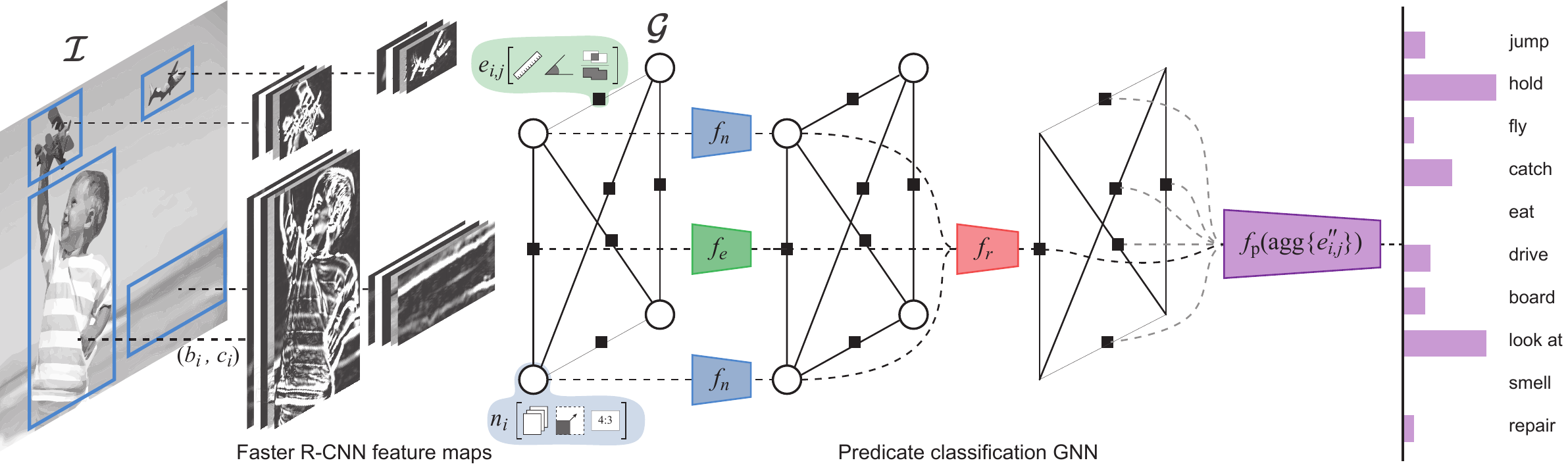}%
    \caption{\small\textbf{A graph neural network (GNN) trained to classify the predicates depicted in a scene.} 
    Object detections extracted through Faster R-CNN are represented as a fully-connected graph.
    The GNN classifier aggregates local information across nodes and produces an image-level predicate prediction.
    The input representation and architecture implicitly encode an inductive bias for pairwise relationships
    }%
    \label{fig:method-predicate-classification}
\end{figure}
Predicate classification as described in \eqref{eq:predicate-classification} is a mapping from image to predicate(s) and as such does not necessarily require an understanding of objects. 
Thus, a simple choice for the classifier would be a convolutional neural network (CNN) trained on image-level predicate labels, \eg ResNeXt \cite{xie2017aggregated}.
However, the raw representation of images as pixels does not explicitly capture the compositional nature of the task.
Instead, we introduce a strong inductive bias towards objects and relationships in both the data representation and the architecture.
Specifically, the module is implemented as a graph neural network (GNN) with architecture similar to \cite{Santoro_NIPS_2017}, that takes as input a graph representation of the image $\Graph = (\Objects, \mathcal{E})$, aggregates information by passing messages over the graph, and produces image-level predicate predictions.
This design choice allows us to later explain the predictions in terms of objects, rather than raw pixels.

Each node in the image graph represents an object $i\in\Objects$ with its spatial and visual features extracted by the object detector, which together we denote as the tuple $\vn_i = (\vn^s_i, \vn^v_i)$.
The image graph is built as fully-connected and therefore impartial to relations between objects.
Directed edges $i\!\rightarrow\!j$ are placed between every pair of nodes, excluding self loops, resulting in $|\Objects|^2 -|\Objects|$ edges.

Node $\vn_i$ and edge $\ve_{i,j}$ representations are first transformed through two small networks $f_n$ and $f_e$:
\begin{align}
    \vn_i'\    &= f_n(\vn_i)      \label{eq:input-node-projection}\\
    \ve_{i,j}' &= f_e(\ve_{i,j}). \label{eq:input-edge-projection}
\end{align}

Then, a relational function $f_r$ aggregates local information by considering pairs of nodes and the edge connecting them:
\begin{align}
    \ve_{i,j}'' &= f_r(\vn_i'\,,\ \ve_{i,j}'\,,\ \vn_j'). \label{eq:relational-fn}
\end{align}
This pairwise function induces an architectural bias towards object-object relationships, which hints at the ultimate goal of relationship detection.

In a fully-supervised scenario, a classification head could be applied to each of the $\ve_{i,j}''$ edges and separate predicate classification losses could be computed using ground-truth pairwise labels $\vp_{i,j}$, \eg~\cite{Qi_ECCV_2018}.
Instead, we consider image-level labels $\vp\in\{0,1\}^K$, where $\evp_k$ indicates the presence of predicate $k$ in the image, \eg $\vp$ would contain $1s$ at the locations of \textit{push, wear, drive} for \figref{fig:intro-relationship-detection}.
Therefore, we aggregate all edge vectors and apply a final prediction function that outputs a binary probability distribution over predicates as in \eqref{eq:predicate-classification}:
\begin{align}
    \vy = f_\text{p} \left(\text{agg} \left\{\ve_{i,j}''\right\} \right) \in \left[0,1\right]^K , \label{eq:aggr-readout-fn}
\end{align}
where \textit{agg} is a permutation-invariant pooling function such as \textit{max}, \textit{sum} or \textit{mean}.

Designed as such, the graph-based predicate classifier can be trained by minimizing the binary cross entropy between predictions and ground-truth labels:
\begin{align}
    \Loss = - \sum_{k=1}^K \left\{\evp_k\log(y_k) + (1-\evp_k)\log(1-y_k)\right\}. \label{eq:pred-class-loss}
\end{align}

\subsection{Explanation-based relationship detection}
\label{sec:method-visual-relation-explanation}
\begin{figure}[th]
    \centering%
    \hfill%
    \begin{subfigure}{.21\textwidth}
        \centering
        \includegraphics[width=\linewidth]{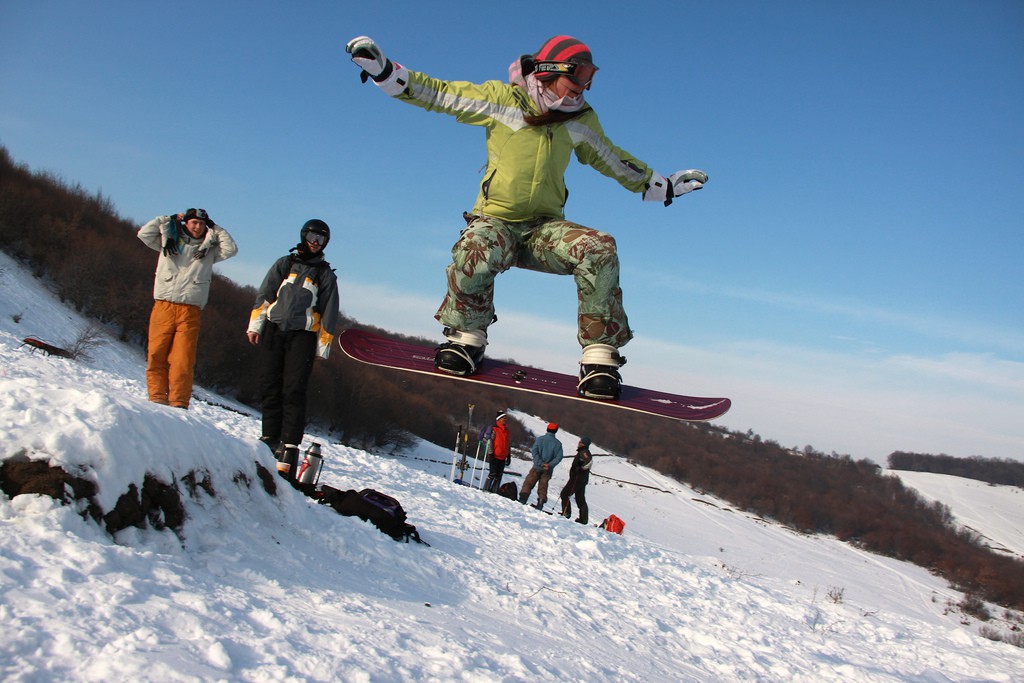}
        \vspace{-.6cm}\caption*{\tiny input}%
    \end{subfigure}\hfill%
    \begin{subfigure}{.21\textwidth}
        \centering
        \includegraphics[width=\linewidth]{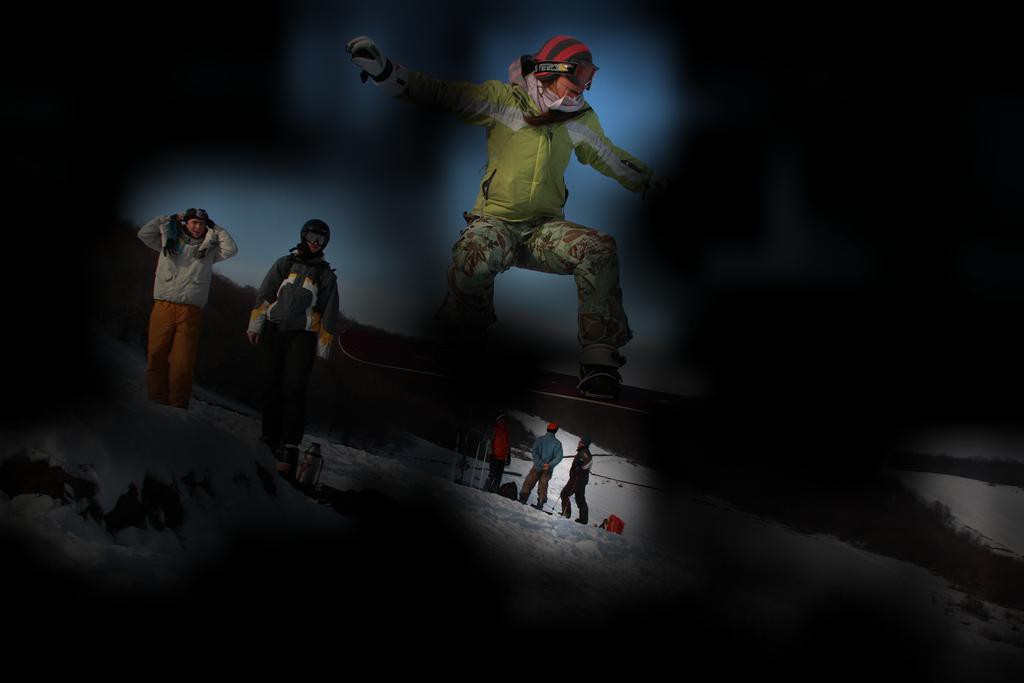}    
        \vspace{-.6cm}\caption*{\tiny wear 74\%}%
    \end{subfigure}\hfill%
    \begin{subfigure}{.21\textwidth}
        \centering
        \includegraphics[width=\linewidth]{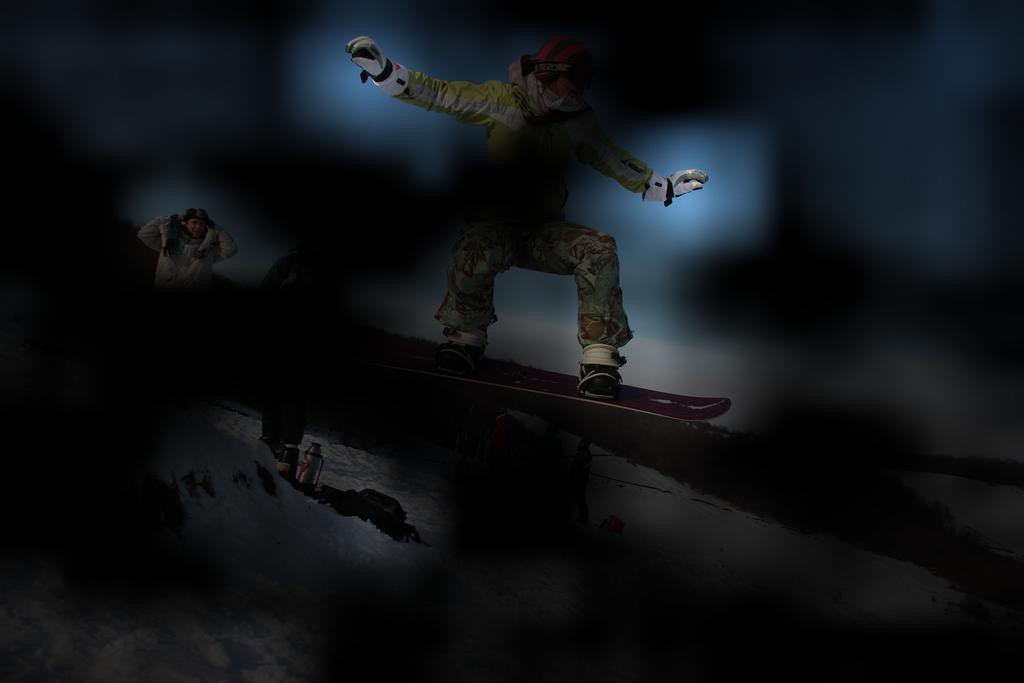}    
        \vspace{-.6cm}\caption*{\tiny above 74\%}%
    \end{subfigure}\hspace*{\fill}%
    \caption{\small\textbf{Grad-CAM heatmap visualization of a ResNet predicate classifier.} Ground-truth annotations contain \textit{person wear jacket} and \textit{person above snowboard}, but it would be hard to identify subjects and objects from the pixel-level explanation.}%
    \label{fig:method-gradcam}%
\end{figure}

Once the predicate classifier is trained, we wish to use it to detect complete relationship triplets $\langle \subj,\pred,\obj \rangle$.
This is where the relational inductive bias introduced for the predicate classifier plays a key role.
In fact, had the predicate classifier been a simple CNN, we would only be able to attribute its predictions to the input pixels, \eg through sensitivity analysis \cite{baehrens2010explain} or Grad-CAM \cite{Selvaraju_ICCV_2017}.
\Figref{fig:method-gradcam} shows an example of Grad-CAM explanations obtained for a ResNeXt architecture \cite{xie2017aggregated} trained for predicate classification on the Visual Relationship Detection dataset~(see \ifappendix\appref{app:resnext-baseline}\else appendix B.3\fi).
While it is possible to guess which areas of the image are relevant for the predicted predicate, it is undoubtedly hard to identify a distinct $(\subj, \obj)$ pair from the pixel-wise heatmaps.

Thanks to the GNN architecture of the previous module, we can instead attribute predicate predictions to the nodes of the input graph, evaluating the importance of \textit{objects} rather than \textit{pixels}. We can then consider all pairs of nodes representing the candidate subject and object of a predicate of interest, score them with a backward explanation procedure and select the top-ranking triplets. 

Specifically, we apply \textit{sensitivity analysis} \cite{baehrens2010explain} to compute the relevance of a node ($r^k_i$) and of an edge ($r^k_{i,j}$) with respect to a predicate $k$:
\begin{align}
    r^k_i &= \norm*{\frac{\partial \evy_k}{\partial \vn_i}}_1 &\text{\footnotesize single-object relevance}\label{eq:relevance-nodes}\\
    r^k_{i,j} &= \norm*{\frac{\partial \evy_k}{\partial \ve_{i,j}}}_1 &\text{\footnotesize object-pair relevance}\label{eq:relevance-edges}
\end{align}
We experimented with different ways to compute these relevances, including $\texttt{gradient}\!\times\!\texttt{input}$, $\texttt{max}(\texttt{gradient}\!\times\!\texttt{input}, 0)$, and the \texttt{L1}, \texttt{L2} norms, but no significant differences were noticed on the validation set. 

The product of these relevances is then used as a proxy for the unnormalized likelihood of a subject-object pair given a predicate (\eqref{eq:visual-relation-explanation}):
\begin{align}
    P(b_\subj=b_i, b_\obj=b_j | k_\pred=k) \propto r^k_i \cdot r^k_{i,j} \cdot r^k_j \label{eq:visual-relation-explanation-approx}.
\end{align}
Rather than computing this quantity for every predicate and for every pair of objects, we limit the search to the $N$ top-scoring predicates, reducing the number of candidates from $K(|\Objects|^2-|\Objects|)$ to $N(|\Objects|^2-|\Objects|)$ relationships. A Big~O complexity that scales as $|\Objects|^2$ might seem unappealing, yet with $|\Objects|\!<\!30$ we could process batches of 128 image graphs in a single pass.
\begin{figure}[!bh]
    \centering%
    \includegraphics[width=.92\textwidth]{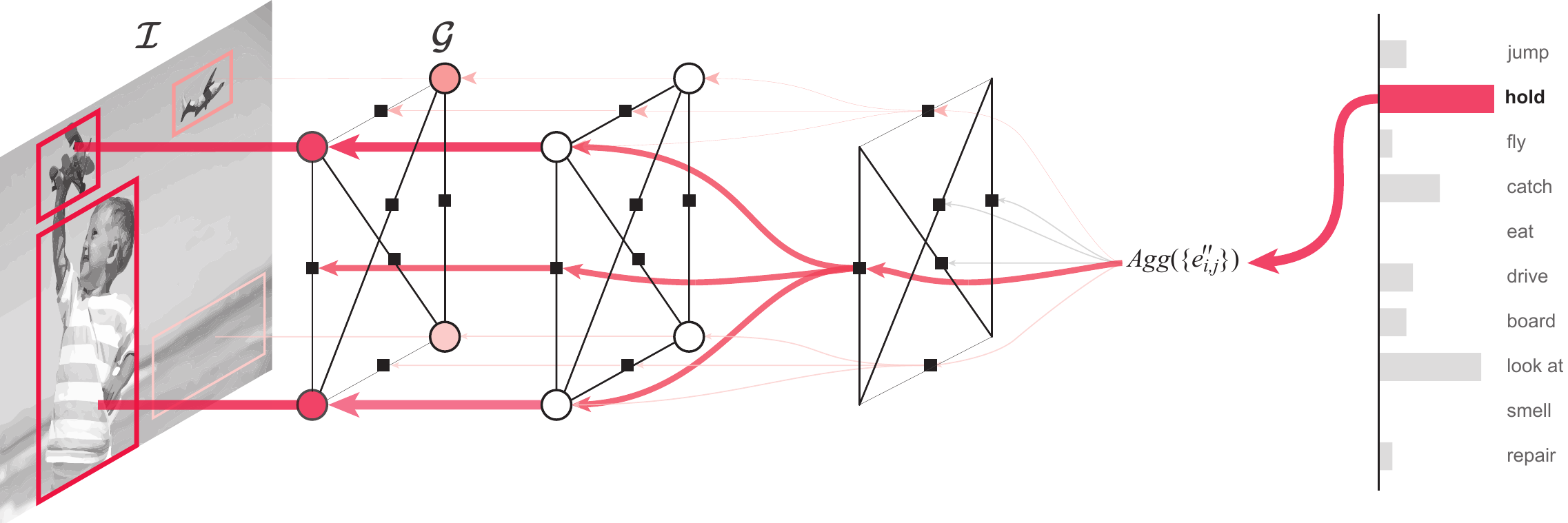}%
    \caption{\small\textbf{Relationship detection through explanation.} A predicate prediction is \textit{explained} by attributing it to the pair of objects in the input that are most relevant for it, effectively recovering a full relationship triplet in the form $\langle\subj,\pred,\obj\rangle$}%
    \label{fig:relationship-explanation}%
\end{figure}
\FloatBarrier
\subsection{Prior over relationships}
\label{sec:method-prior-over-relations}
Learning to detect $\langle\subj,\pred,\obj\rangle$ relations using image-level predicate labels is inherently ill-defined.
Consider the task of learning a new predicate, \eg \textit{squanch}.
By observing a sufficient number of labeled images, we could learn that two specific objects are often in a \textit{squanch} relationship.
However, we would not be able to determine which should be the subject and which the object, \ie the direction of such relation, without semantic knowledge about the new word (can things be \textit{squanchier than} others? can objects \textit{squanch} each other?).

\Eqref{eq:prior-over-relations} represents the belief over which categories can act as subject and objects of a certain predicate.
In fully- or weakly-supervised scenarios, where $\langle\subj,\pred,\obj\rangle$ triplets are available during training, a relationship detector would learn such biases directly from data.
Our graph-based predicate classifier, trained only image-level predicate annotations, can indeed learn to recognize object-object relations and to assign high probability to meaningful pairs (\eqref{eq:visual-relation-explanation-approx}), but neither the training signal nor the inductive biases contain hints about directionality.
In fact, the relevance $r^k_{i,j}$ is in no way constrained to represent the relationship that has $i$ as subject and $j$ as object, even though \eqref{eq:relational-fn} considers the edge $i\!\rightarrow\!j$.
Thus the explanation (\eqref{eq:visual-relation-explanation-approx}) for \textit{hold} might score both \textit{person~hold~pencil} and its semantic opposite \textit{pencil~hold~person} equally.

Previous work \cite{Lu_ECCV_2016} use \texttt{word2vec} \cite{mikolov2013efficient} embeddings of $\langle\subj,\pred,\obj\rangle$ triplets from the training set to form a semantically-grounded prior.
Instead, we compute a simple frequency-based prior $\text{freq}(c_i,c_j|k)$ over a small validation set, to avoid including exclusive relationship information from the training set (app. \ifappendix\ref{app:relationship-prior}\else C.3\fi).

\section{Experiments}
\label{sec:experiments}

In this section, we test our weakly-supervised method for visual relationship detection on three different datasets, each one presenting specific challenges and different evaluation metrics.
Before discussing the individual experiments, we provide further implementation details about the object detection, predicate classification and visual relationship explainer modules.
Additional experiments and ablation studies can be found in \ifappendix\appref{app:additional-experiments}\else appendix C\fi.

\subsection{Setup}
\textbf{Object detection.} Our object detection module is based on the \texttt{detectron2} \cite{wu2019detectron2} implementation of Faster R-CNN~\cite{ren2015faster}.
Given an object $i$ and its bounding box $b_i$, either from the ground-truth annotations or detected by Faster R-CNN, we use $\textsc{RoIAlign}$~\cite{he2017mask} to pool a $256 \times\!7\!\times\!7$ feature volume $\vn^v_i$ from the pyramid of features~\cite{lin2017feature} built on top of a ResNeXt-101 backbone~\cite{xie2017aggregated}. 
Furthermore, we compute a feature vector $\vn^s_i$ that represents the spatial configuration of $b_i$.
Specifically, the tuple of spatial and visual features $\vn_i = (\vn^s_i, \vn^v_i)$ is defined as:
\begin{align}
    \vn^s_i &= \left[ \frac{w_i}{h_i},\ \frac{h_i}{w_i},\ \frac{w_i h_i}{W H} \right] &\text{\footnotesize spatial features} \label{eq:spatial-features} \\
    \vn^v_i &= \textsc{RoiAlign}\left(\textsc{FPN}(\Image), b_i\right), &\text{\footnotesize visual features} \label{eq:visual-features}
\end{align}
where ($w_i$, $h_i$) and ($W$, $H$) represent width and height of the box $b_i$ and of the image $\Image$ respectively, $\textsc{FPN}$ is the feature pyramid network used to extract visual features from the whole image, and $\textsc{RoIAlign}$ is the pooling operation applied to the feature pyramid to extract features relative to the box $b_i$.

Edge attributes $\ve_{i,j}$ are chosen to represent the spatial configuration of the pair of objects they connect:
\begin{align}
    \ve_{i,j} = \left[ \frac{\norm{\vx_j - \vx_i}}{\sqrt{WH}} ,\ \sin(\angle_{ij}),\ \cos(\angle_{ij}),\ \IoU(b_j, b_i),\ \frac{w_j h_j}{w_i h_i} \right],
\end{align}
where $\vx_i\in\mathbb{R}^2_{+}$ is the center of $b_i$, $\angle_{ij}$ is the angle between $\vx_j - \vx_i$ and the positive horizontal axis, and $\IoU$ is the intersection over union of the two boxes.

\noindent \textbf{Predicate classifier.} 
At training time, the input of the predicate classifier described in sec. \ref{sec:method-predicate-classification} is a fully-connected graph of ground-truth objects.
During inference, we apply the object detector and build a graph with all objects having confidence score of 30\% or more.
For each dataset, the hyperparameters of the GNN-based predicate classifier are selected on a validation split of 15\% training images.
The following values apply to the HICO-DET dataset, more details about the hyperparameter space are available in \ifappendix\appref{app:predicate-classifier-hyperparameters}\else appendix B.2\fi.

The input node function $f_n$ is implemented as i) a $2\!\times\!(\textsc{Conv}+\ReLU)$ network applied to $\vn^v$, where the convolutional layers employ 256 kernels of size $3\!\times\!3$ each, and ii) a $\textsc{Linear}+\ReLU$ operation that transforms $\vn^s$ into a 1024-vector.
The input edge functions $f_e$ consist of a $\textsc{Linear}+\ReLU$ operation that outputs a 1024-vector of transformed edge features.
The relational function $f_r$ in \eqref{eq:relational-fn} is implemented as a $\textsc{Linear}+\ReLU$ operation where the features of two nodes and of the directed edge between them are concatenated at the input.
The output of $f_r$ is a 1024-vector for each ordered pair of nodes.
For all datasets, the aggregation function in \eqref{eq:aggr-readout-fn} is element-wise \textit{max}, and $f_p$ is a $\textsc{Linear}+\textsc{Sigmoid}$ operation that outputs a $K-$vector of binary probabilities.

We train the weights of the predicate classifier by minimizing the loss in \eqref{eq:pred-class-loss} with the Adam optimizer~\cite{kingma2014adam} with $10^{-3}$ initial learning rate and $10^{-5}$ weight decay.
During training, we track \texttt{recall@5}, \ie the fraction of ground-truth predicates retrieved among the top-5 confident predictions for an image.
We let the optimization run on batches of 128 graphs for 18 epochs, at which point the classifier achieves 94\% recall on a validation split.

\noindent \textbf{Relationship detector.} 
The explanation-based relationship detection algorithm described in \secref{sec:method-visual-relation-explanation} does not have many hyperparameters.
We tried i) whether to multiply the gradient with the input when computing relevances, ii) which norm to use between L1, L2 and max(L1,0), and iii) the number $N$ of top-scoring predicates whose gradient is traced back to the input to identify relevant triplets. 
As observed in \cite{choe2020evaluating}, optimizing these parameters on the whole training set would violate the premise of weakly-supervised learning by accessing fully-labeled data.
Therefore, we employ once again the 15\% validation split used to optimize the classifier, assuming that in a real-world scenario it should always be possible to manually annotate a small subset of images for validation purposes.
The best choice of $N$ for all datasets was found to be 10, while the other two parameters seem to have little effect on performance.

\subsection{HICO-DET}
\label{sec:experiments-hico}
The Humans Interacting with Common Objects (HICO-DET) dataset contains $\sim\!50K$ exhaustively annotated images of \textit{human-object interactions} (HOI), split into $\sim\!40K$ train and $\sim\!10K$ test images \cite{Chao_ICCV_2015,Chao_WACV_2018}.
The subject of a relationship is always \textit{person}, the 117 predicates cover a variety of human-centric actions (\eg \textit{cook, wash, paint}), and the 80 objects categories are those defined as \texttt{thing\_classes} in MS-COCO~\cite{lin2014microsoft}. 
We can therefore use the pre-trained object detector from~\cite{wu2019detectron2}, of which we report performances in \ifappendix\appref{app:datasets-hico}\else appendix A.1\fi.

The nature of this dataset allows us to embed the relationship prior in the graph itself. 
A fully-connected graph encodes a uniform prior, \ie no preference about subject-object pairs, while a sparse graph containing only edges from humans to objects encodes a bias towards \textit{human-object interactions}.

The metric for this dataset is the 11-point interpolated mean Average Precision (mAP) \cite{everingham2010pascal} computed over the 600 human-object interaction classes of the dataset \cite{Chao_WACV_2018}. 
The following criteria should be met for a detected triplet to match with a ground-truth triplet: a) subject, predicate and object categories match, and b) subject boxes overlap with $\IoU\!>\!.5$, and c) object boxes overlap with $\IoU\!>\!.5$, and d) the ground-truth triplet was not matched with a previously-considered detected triplet.
\begin{figure}[!b]
    \centering%
    \captionsetup[subfigure]{justification=centering}%
    \begin{subfigure}[t]{.23\linewidth}%
        \centering%
        \includegraphics[width=\linewidth]{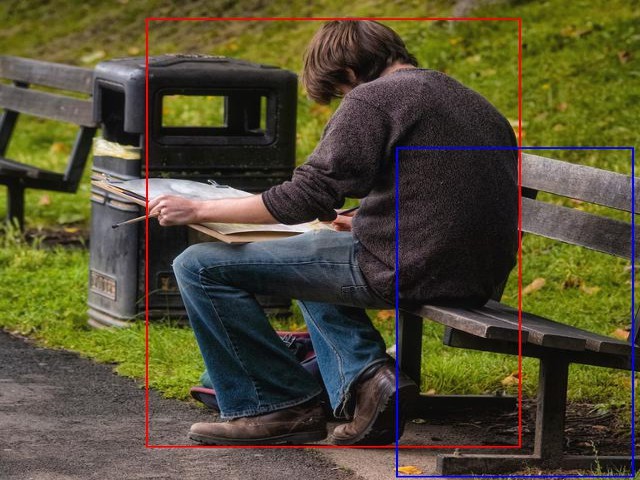}
        \vspace{-.6cm}\caption*{\tiny p. sit on bench}%
    \end{subfigure}\hfill%
    \begin{subfigure}[t]{.23\linewidth}%
        \centering%
        \includegraphics[width=\linewidth]{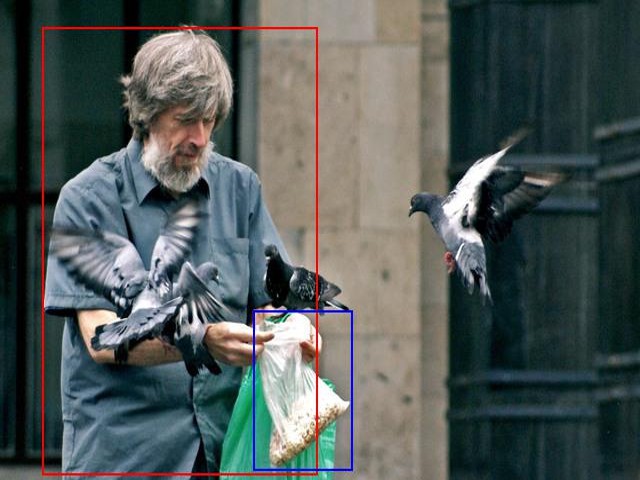}
        \vspace{-.6cm}\caption*{\tiny p. carry handbag}%
    \end{subfigure}\hfill%
    \begin{subfigure}[t]{.23\linewidth}%
        \centering%
        \includegraphics[width=\linewidth]{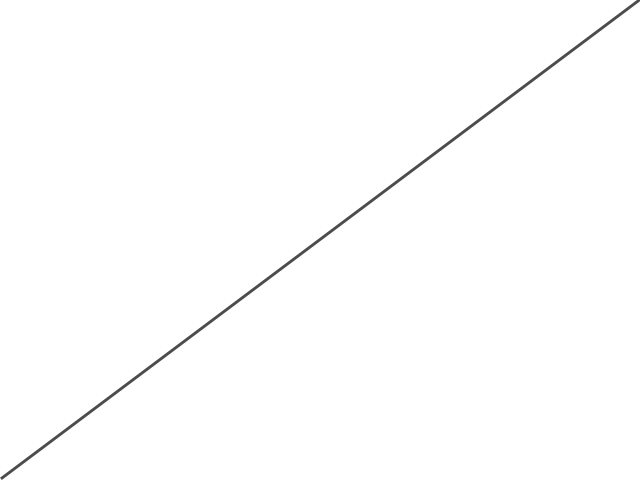}
        \vspace{-.6cm}\caption*{\tiny (not possible with ground-truth objects)}%
    \end{subfigure}\hfill%
    \begin{subfigure}[t]{.23\linewidth}%
        \centering%
        \includegraphics[width=\linewidth]{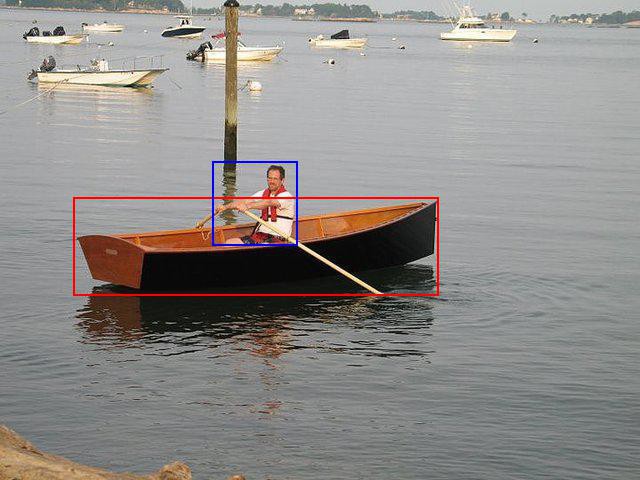}
        \vspace{-.6cm}\caption*{\tiny boat row person \\(subj-obj inversion)}%
    \end{subfigure}\\%
    \begin{subfigure}[t]{.23\linewidth}%
        \centering%
        \includegraphics[width=\linewidth]{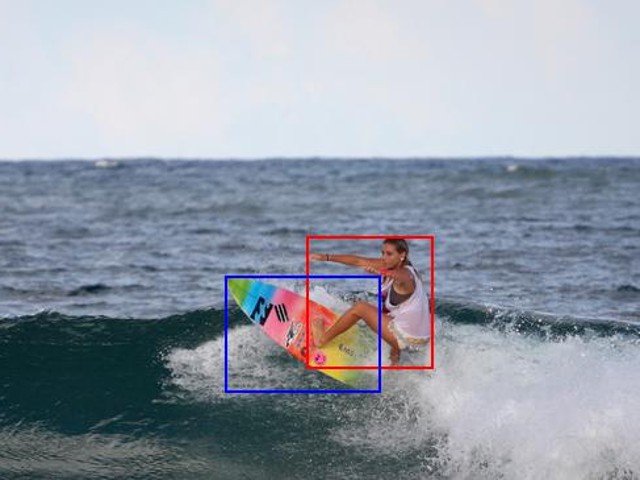}
        \vspace{-.6cm}\caption*{\tiny p. ride surfboard}%
    \end{subfigure}\hfill%
    \begin{subfigure}[t]{.23\linewidth}%
        \centering%
        \includegraphics[width=\linewidth]{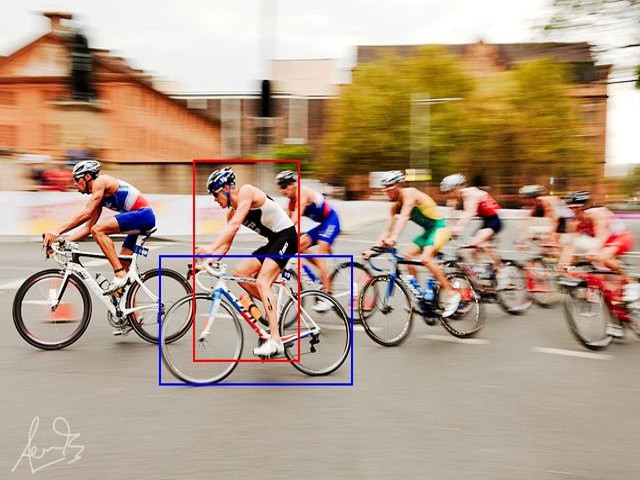}
        \vspace{-.6cm}\caption*{\tiny p. straddle bicycle}%
    \end{subfigure}\hfill%
    \begin{subfigure}[t]{.23\linewidth}%
        \centering%
        \includegraphics[width=\linewidth]{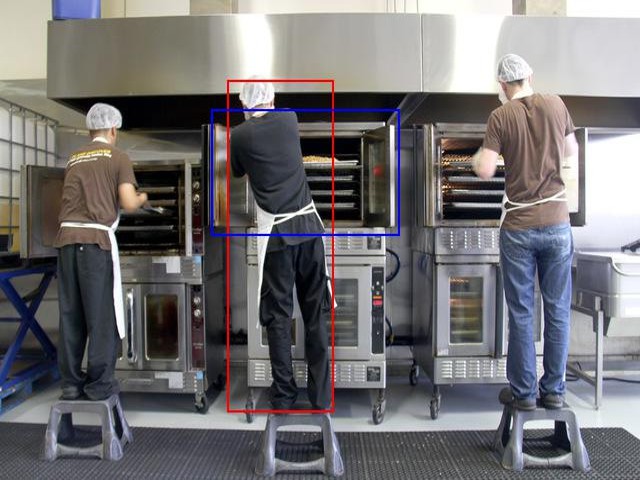}
        \vspace{-.6cm}\caption*{\tiny p. operate microwave \\(p. operate oven)}%
    \end{subfigure}\hfill%
    \begin{subfigure}[t]{.23\linewidth}%
        \centering%
        \includegraphics[width=\linewidth]{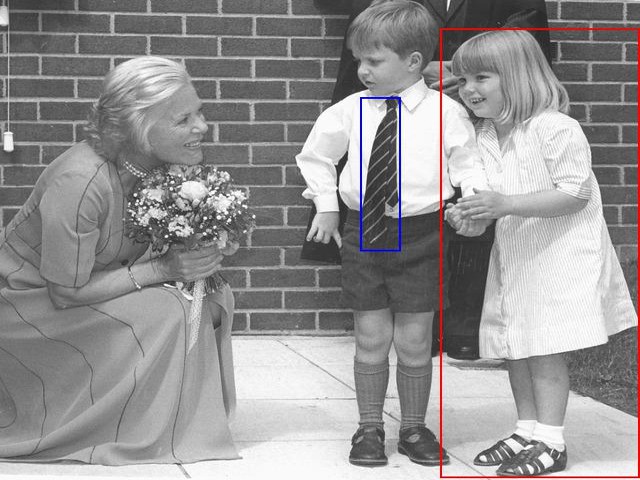}
        \vspace{-.6cm}\caption*{\tiny person wear tie \\(wrong subj-obj pair)}%
    \end{subfigure}%
    \caption{\small \textbf{Relationship detection on HICO-DET.} Top row uses GT objects, bottom row uses Faster R-CNN objects. Left to right: correct relationship detection, correct but missing ground-truth, incorrect due to object misdetection, incorrect detection (selected predictions of our model using a uniform relationship prior)}%
    \label{fig:experiments-hico}%
\end{figure}
\Tabref{tab:experiments-hico-map} reports mAP for the standard splits of HICO-DET\cite{Chao_WACV_2018}: all 600 human-object interactions, 138 rare triplets, and 462 non-rare triplets~(10 or more training samples).
We compare with various fully-supervised baselines including the original HO-RCNN from \cite{Chao_WACV_2018} and the method from \cite{Peyre_ICCV_2019} that uses semantic and visual analogies to improve detection of rare and unseen triplets. 
Despite the weaker supervision signal, the strong inductive bias towards pairwise relationships allows our explanation method to achieve higher mAP for both the uniform and human-object priors.
\begin{table}[!t]
    \centering%
    \caption{\small \textbf{Mean Average Precision on the HICO-DET dataset.} The choice of relationship prior embedded in the graph is indicated in parentheses}%
    \label{tab:experiments-hico-map}%
    \scriptsize%
    \begin{tabular}{@{}rrrr@{}}
\toprule
\multicolumn{1}{c}{}                  & \multicolumn{1}{c}{Full (600)} & \multicolumn{1}{c}{Rare (138)} & \multicolumn{1}{c}{Non-rare (462)} \\ \midrule
\textbf{Fully supervised}             &                                &                                &                                    \\
Chao \cite{Chao_WACV_2018}            & 7.81                           & 5.37                           & 8.54                               \\
InteractNet \cite{Gkioxari_CVPR_2018} & 9.94                           & 7.16                           & 10.77                              \\
GPNN \cite{Qi_ECCV_2018}              & 13.11                          & 9.34                           & 14.23                              \\
iCAN \cite{ican_BMVC_2018}            & 14.84                          & 10.45                          & 16.15                              \\
Analogies \cite{Peyre_ICCV_2019}      & 19.40                          & 14.60                          & 20.90                              \\
\textbf{Weakly supervised}            &                                &                                &                                    \\
Ours (uniform)                        & 24.25                          & 20.23                          & 25.45                              \\
Ours (human-object)                   & 28.04                          & 24.63                          & 29.06                              \\ \bottomrule
\end{tabular}
\end{table}%
\FloatBarrier%
\subsection{Visual Relationship Detection dataset}
\label{sec:experiments-vrd}
The Visual Relationship Detection dataset (VRD) contains $\sim\!5000$ annotated images, split into $\sim\!4000$ train and $\sim\!1000$ test images~\cite{Johnson_CVPR_2015,Lu_ECCV_2016}.
The 70 predicates in this dataset include both verbs and spatial relationships, \eg \textit{carry}, \textit{next to}.
The 100 object categories cover both well spatially-defined objects such as \textit{bottle} and concepts like \textit{sky} and \textit{road}, that are harder to localize.
For this set of objects there is no ready-to-use object detector, therefore we finetune a \texttt{detectron2} model using annotations from the training set (details in \ifappendix\appref{app:datasets-vrd}\else appendix A.2\fi).

The standard metric for VRD~\cite{Lu_ECCV_2016} is \texttt{recall@x} \ie fraction of ground-truth triplets retrieved among the $x$ top-ranked detections \cite{alexe2012measuring}.
Here, recall is preferred over mAP since it does not penalize the retrieval of triplets that exist in the image, but are missing in the ground-truth.
Criteria for true positive in VRD follow those of HICO-DET, and are used in the following settings~\cite{Lu_ECCV_2016}:
\begin{description}
    \item[Predicate detection:] objects for the image graph come from ground-truth annotations,allowing to test the explanation-based retrieval of relationships under perfect object detection conditions (classification and localization).
    \item[Phrase detection:] objects come from Faster R-CNN proposals, but $\IoU\!>\!.5$ is evaluated on the union box of subject and object, effectively localizing the entire relationship as a single image region, or \textit{visual phrase} \cite{Sadeghi_CVPR_2011}.
    \item[Relationship detection:] objects come from Faster R-CNN proposals, subject and object boxes are required to individually overlap with their corresponding boxes in the ground-truth (same as HICO-DET).
\end{description}

\noindent As shown in \tabref{tab:experiments-vrd-recall}, our method achieves recall scores R@100 close to a fully-supervised baseline \cite{Lu_ECCV_2016}, despite the weaker training signal.
By analyizing the top 100 predictions of a model with uniform prior, we often observed the coappearance of a relationship and its semantic opposite, \eg~\textit{person drive car} and \textit{car drive person}, which possibly ``wastes'' half of the top-x detection due to incorrect directionality (corroborated by the gap between R@50 and R@100 of ours-uniform).
Importantly, moving from a uniform to a frequency-based prior almost doubles R@50, which highlights the importance of the relationship prior in connection with our method. We expect that including a stronger prior, \eg based on natural-language embeddings of objects and predicates, would further improve detection of semantically-correct relationships.
\begin{table}[!t]
    \centering%
    \caption{\small \textbf{Recall at 50 and 100 on the VRD dataset.} Comparison of fully- and weakly-supervised methods. The choice of relationship prior is indicated in parentheses}%
    \label{tab:experiments-vrd-recall}%
    \scriptsize%
    \begin{tabular}{@{}rrrrrrr@{}}
\toprule
\multicolumn{1}{c}{}                    & \multicolumn{2}{c}{GT objects}                       & \multicolumn{4}{c}{R-CNN objects}                                                                           \\
\multicolumn{1}{c}{}                    & \multicolumn{2}{c}{Predicate det.}                   & \multicolumn{2}{c}{Phrase det.}                      & \multicolumn{2}{c}{Relation. det.}                \\
\multicolumn{1}{c}{}                    & \multicolumn{1}{c}{R@50} & \multicolumn{1}{c}{R@100} & \multicolumn{1}{c}{R@50} & \multicolumn{1}{c}{R@100} & \multicolumn{1}{c}{R@50} & \multicolumn{1}{c}{R@100} \\ \midrule
\textbf{Fully supervised}               &                          &                           &                          &                           &                          &                           \\
Visual Phrases \cite{Sadeghi_CVPR_2011} & 0.9                      & 1.9                       & 0.04                     & 0.07                      & -                        & -                         \\
Visual \cite{Lu_ECCV_2016}              & 3.5                      & 3.5                       & 0.7                      & 0.8                       & 1.0                      & 1.1                       \\
Visual+Language \cite{Lu_ECCV_2016}     & 47.9                     & 47.9                      & 16.2                     & 17.0                      & 13.9                     & 14.7                      \\
Sup. PPR-FCN \cite{Zhang_ICCV_2017}     & 47.4                     & 47.4                      & 19.6                     & 23.2                      & 14.4                     & 15.7                      \\
Peyre \cite{Peyre_ICCV_2017}            & 52.6                     & 52.6                      & 17.9                     & 19.5                      & 15.8                     & 17.1                      \\
\textbf{Weakly sup. (subj,pred,obj)}    &                          &                           &                          &                           &                          &                           \\
PPR-FCN \cite{Zhang_ICCV_2017}          & -                        & -                         & 6.9                      & 8.2                       & 5.9                      & 6.3                       \\
Peyre \cite{Peyre_ICCV_2017}            & 46.8                     & 46.8                      & 16.0                     & 17.4                      & 14.1                     & 15.3                      \\
\textbf{Weakly sup. (pred only)}        &                          &                           &                          &                           &                          &                           \\
Ours (uniform)                          & 27.3                     & 47.1                      & 6.8                      & 13.0                      & 5.3                      & 8.4                       \\
Ours (frequentist)                      & 43.0                     & 57.4                      & 14.8                     & 20.2                      & 10.6                     & 13.2                      \\ \bottomrule
\end{tabular}

\end{table}

The test set of VRD contains a triplets that never occur during training and can be used to evaluate zero-shot generalization.
As shown in \tabref{tab:experiments-vrd-recall-zero-shot}, our method performs on a par with other methods that use stronger annotations and explicitly improve generalization through language embeddings \cite{Lu_ECCV_2016} or visual analogy transformations \cite{Peyre_ICCV_2019}.
Expectedly, the freq-based prior computed on the validation set does not improve recall of unseen triplets.
To verify importance of this term, we show that a simple prior with access to a few zero-shot triplets readily improves recall.
Clearly, peeking at the test set is not correct practice, but serves as a proxy for what could be achieved by improving this term, \eg via incorporating language or visual analogies.
The next experiment better demonstrates the generalization of our method to unseen triplets.

\subsection{Unusual Relations dataset}
\label{sec:experiments-unrel}
The Unusual Relations dataset (UnRel) is an evaluation-only collection of $\sim\!1000$ images, which shares the same vocabulary as VRD and depicts rarely-occurring relationships \cite{Peyre_ICCV_2017}.
For relationship detection methods trained on $\langle\subj,\pred,\obj\rangle$ triplets, this dataset represents a benchmark for zero-shot retrieval of triplets not seen during training.
E.g. our predicate classifier trained on VRD has clearly encountered \textit{hold} during training, but never in \textit{person hold plane}~(\figref{fig:method-predicate-classification}).

In \tabref{tab:experiments-unrel-map} we report mAP over the 76 unusual triplets of UnRel.
We follow the evaluation setup of \cite{Peyre_ICCV_2017}: the test set of VRD is mixed in to act as distractor, up to 500 candidate triplets per image are retained, and they are matched if $\IoU\!>\!.3$.
Since the average number of detected objects per image is small, $\sim\!4$, we increase the number of top-scoring predicates considered in the explanation step to $N=50$.
Differently from \cite{Peyre_ICCV_2017,Peyre_ICCV_2019}, we use obj. detection scores for ranking triplets, and we do not introduce a \textit{no-interaction} predicate.
Compared to recall, mAP is less affected by unseen triplets and the prior from VRD results effective.

\begin{table}[!t]
    \centering%
    \caption{\small\textbf{Zero-shot recall on the VRD dataset:} triplets from the test set that are never seen during training. The choice of relationship prior is indicated in parentheses}%
    \label{tab:experiments-vrd-recall-zero-shot}%
    \scriptsize%
    \begin{tabular}{@{}rrrrrrr@{}}
\toprule
\multicolumn{1}{c}{}                 & \multicolumn{2}{c}{GT objects}                       & \multicolumn{4}{c}{R-CNN objects}                                                                           \\
\multicolumn{1}{c}{}                 & \multicolumn{2}{c}{Predicate det.}                   & \multicolumn{2}{c}{Phrase det.}                      & \multicolumn{2}{c}{Relation. det.}                \\
\multicolumn{1}{c}{}                 & \multicolumn{1}{c}{R@50} & \multicolumn{1}{c}{R@100} & \multicolumn{1}{c}{R@50} & \multicolumn{1}{c}{R@100} & \multicolumn{1}{c}{R@50} & \multicolumn{1}{c}{R@100} \\ \midrule
\textbf{Fully supervised}            &                          &                           &                          &                           &                          &                           \\
Visual \cite{Lu_ECCV_2016}           & 3.5                      & 3.5                       & 0.7                      & 0.8                       & 1.0                      & 1.1                       \\
Visual+Language \cite{Lu_ECCV_2016}  & 8.5                      & 8.5                       & 3.4                      & 3.8                       & 3.1                      & 3.5                       \\
Peyre 2017 \cite{Peyre_ICCV_2017}         & 21.6                     & 21.6                      & 6.8                      & 7.8                       & 6.4                      & 7.4                       \\
\textbf{Weakly sup. (subj,pred,obj)} &                          &                           &                          &                           &                          &                           \\
Peyre 2017 \cite{Peyre_ICCV_2017}         & 19.0                     & 19.0                      & 6.9                      & 7.4                       & 6.7                      & 7.1                       \\
\textbf{Weakly sup. (pred only)}     &                          &                           &                          &                           &                          &                           \\
Ours (uniform)                       & 13.7                     & 29.2                      & 3.8                      & 6.5                       & 2.8                      & 4.6                       \\
Ours (VRD freq.)                     & 13.5                     & 28.2                      & 4.4                      & 6.4                       & 3.3                      & 4.6                       \\
Ours (Zero freq.)                    & 20.5                     & 37.0                      & 4.7                      & 8.2                       & 4.0                      & 6.4                       \\ \bottomrule
\end{tabular}

\end{table}

\begin{table}[t]
    \centering%
    \caption{\small\textbf{Mean Average Precision on UnRel with VRD as a distractor}}%
    \label{tab:experiments-unrel-map}%
    \scriptsize%
    \begin{tabular}{@{}rcccc@{}}
\toprule
                                        & GT objects & \multicolumn{3}{c}{R-CNN objects}  \\
                                        & Predicate  & Phrase & Subj. only & Relationship \\ \midrule
\textbf{Fully supervised}            &            &        &            &              \\
Peyre 2017 \cite{Peyre_ICCV_2017}    & 62.6       & 14.1   & 12.1       & 9.9          \\
Analogies \cite{Peyre_ICCV_2019}     & 63.9       & 17.5   & 15.9       & 13.4         \\
\textbf{Weakly sup. (subj,pred,obj)} &            &        &            &              \\
Peyre 2017 \cite{Peyre_ICCV_2017}    & 58.5       & 13.4   & 11.0       & 8.7          \\
\textbf{Weakly sup. (pred only)}     &            &        &            &              \\
Ours (uniform)                       & 70.9       & 19.8   & 18.1       & 14.9         \\
Ours (frequency)                     & 70.6       & 20.0   & 18.3       & 15.1         \\ \bottomrule
\end{tabular}

\end{table}

\section{Conclusion}
\label{sec:conclusion}
We considered the task of learning visual relationship detection with weak image-level predicate labels. While this makes learning significantly harder, it enables collecting datasets that are more representative of possible relations without suffering from combinatorial scaling of search queries and annotation cost. 

Using pretrained object detectors, strong inductive bias via graph networks, backward explanations, and a direction prior, we showed that it is possible to achieve results on par with recent works that benefit from stronger supervision.

An issue with predicate-only annotation is the lack of directional information, which can only be provided using auxiliary sources such as language. While we mitigated this issue through a simple frequentist prior, an important future direction is to solve it in a principled way. For instance, one can annotate a subset of images with unlocalized image-level triplets, only to disambiguate the direction of the relations. Note that, since such a dataset does not have to be exhaustively annotated for all triplets, the collection cost would be negligible.

Finally, another interesting direction is to study the proposed explanation-based weakly-supervised method in other domains such as situation recognition~\cite{Li_ICCV_2017}, video recognition~\cite{Yan_AAAI_2018}, segmentation~\cite{Qi_ICCV_2017}, chemistry~\cite{Do_KDD} and biology~\cite{Tsubaki_BioInf_2019}.

\noindent\textbf{Acknowledgements.}
Funded by Swedish Research Council project 2017-04609 and by Wallenberg AI, Autonomous Systems and Software Program (WASP).

\bibliographystyle{splncs04}
\bibliography{main}

\begin{thebibliography}{10}
\providecommand{\url}[1]{\texttt{#1}}
\providecommand{\urlprefix}{URL }
\providecommand{\doi}[1]{https://doi.org/#1}

\bibitem{alexe2012measuring}
Alexe, B., Deselaers, T., Ferrari, V.: Measuring the objectness of image
  windows. IEEE transactions on pattern analysis and machine intelligence
  \textbf{34}(11),  2189--2202 (2012)

\bibitem{Bach_Plos_2015}
Bach, S., Binder, A., Montavon, G., Klauschen, F., M{\"u}ller, K.R., Samek, W.:
  On pixel-wise explanations for non-linear classifier decisions by layer-wise
  relevance propagation. PloS one  \textbf{10}(7) (2015)

\bibitem{baehrens2010explain}
Baehrens, D., Schroeter, T., Harmeling, S., Kawanabe, M., Hansen, K.,
  M{\~A}{\v{z}}ller, K.R.: How to explain individual classification decisions.
  Journal of Machine Learning Research  \textbf{11}(Jun),  1803--1831 (2010)

\bibitem{Baldassarre_ICMLW_2019}
Baldassarre, F., Azizpour, H.: Explainability techniques for graph
  convolutional networks. In: International Conference on Machine Learning
  (ICML) Workshops, 2019 Workshop on Learning and Reasoning with
  Graph-Structured Representations (2019)

\bibitem{Bilen_CVPR_2016}
Bilen, H., Vedaldi, A.: Weakly supervised deep detection networks. In: The IEEE
  Conference on Computer Vision and Pattern Recognition (CVPR) (June 2016)

\bibitem{Chao_WACV_2018}
Chao, Y.W., Liu, Y., Liu, X., Zeng, H., Deng, J.: Learning to detect
  human-object interactions. In: 2018 ieee winter conference on applications of
  computer vision (wacv). pp. 381--389. IEEE (2018)

\bibitem{Chao_ICCV_2015}
Chao, Y.W., Wang, Z., He, Y., Wang, J., Deng, J.: Hico: A benchmark for
  recognizing human-object interactions in images. In: Proceedings of the IEEE
  International Conference on Computer Vision. pp. 1017--1025 (2015)

\bibitem{choe2020evaluating}
Choe, J., Oh, S.J., Lee, S., Chun, S., Akata, Z., Shim, H.: Evaluating weakly
  supervised object localization methods right. arXiv preprint arXiv:2001.07437
   (2020)

\bibitem{Do_KDD}
Do, K., Tran, T., Venkatesh, S.: Graph transformation policy network for
  chemical reaction prediction. In: Proceedings of the 25th ACM SIGKDD
  International Conference on Knowledge Discovery \& Data Mining. pp. 750--760
  (2019)

\bibitem{Durand_CVPR_2017}
Durand, T., Mordan, T., Thome, N., Cord, M.: Wildcat: Weakly supervised
  learning of deep convnets for image classification, pointwise localization
  and segmentation. In: Proceedings of the IEEE conference on computer vision
  and pattern recognition. pp. 642--651 (2017)

\bibitem{everingham2010pascal}
Everingham, M., Van~Gool, L., Williams, C.K., Winn, J., Zisserman, A.: The
  pascal visual object classes (voc) challenge. International journal of
  computer vision  \textbf{88}(2),  303--338 (2010)

\bibitem{Galleguillos_CVPR_2008}
Galleguillos, C., Rabinovich, A., Belongie, S.: Object categorization using
  co-occurrence, location and appearance. In: 2008 IEEE Conference on Computer
  Vision and Pattern Recognition. pp.~1--8. IEEE (2008)

\bibitem{ican_BMVC_2018}
Gao, C., Zou, Y., Huang, J.B.: ican: Instance-centric attention network for
  human-object interaction detection  (2018)

\bibitem{Ge_ICCV_2019}
Ge, W., Guo, S., Huang, W., Scott, M.R.: Label-penet: Sequential label
  propagation and enhancement networks for weakly supervised instance
  segmentation. In: Proceedings of the IEEE International Conference on
  Computer Vision. pp. 3345--3354 (2019)

\bibitem{Gkioxari_CVPR_2018}
Gkioxari, G., Girshick, R., Doll{\'a}r, P., He, K.: Detecting and recognizing
  human-object interactions. In: Proceedings of the IEEE Conference on Computer
  Vision and Pattern Recognition. pp. 8359--8367 (2018)

\bibitem{he2017mask}
He, K., Gkioxari, G., Doll{\'a}r, P., Girshick, R.: Mask r-cnn. In: Proceedings
  of the IEEE international conference on computer vision. pp. 2961--2969
  (2017)

\bibitem{Johnson_CVPR_2017}
Johnson, J., Hariharan, B., van~der Maaten, L., Fei-Fei, L., Lawrence~Zitnick,
  C., Girshick, R.: Clevr: A diagnostic dataset for compositional language and
  elementary visual reasoning. In: Proceedings of the IEEE Conference on
  Computer Vision and Pattern Recognition. pp. 2901--2910 (2017)

\bibitem{Johnson_CVPR_2015}
Johnson, J., Krishna, R., Stark, M., Li, L.J., Shamma, D., Bernstein, M.,
  Fei-Fei, L.: Image retrieval using scene graphs. In: Proceedings of the IEEE
  conference on computer vision and pattern recognition. pp. 3668--3678 (2015)

\bibitem{kingma2014adam}
Kingma, D.P., Ba, J.: Adam: A method for stochastic optimization. arXiv
  preprint arXiv:1412.6980  (2014)

\bibitem{Lee_CVPR_2019}
Lee, J., Kim, E., Lee, S., Lee, J., Yoon, S.: Ficklenet: Weakly and
  semi-supervised semantic image segmentation using stochastic inference. In:
  Proceedings of the IEEE conference on computer vision and pattern
  recognition. pp. 5267--5276 (2019)

\bibitem{Li_ICCV_2017}
Li, R., Tapaswi, M., Liao, R., Jia, J., Urtasun, R., Fidler, S.: Situation
  recognition with graph neural networks. In: Proceedings of the IEEE
  International Conference on Computer Vision. pp. 4173--4182 (2017)

\bibitem{lin2017feature}
Lin, T.Y., Doll{\'a}r, P., Girshick, R., He, K., Hariharan, B., Belongie, S.:
  Feature pyramid networks for object detection. In: Proceedings of the IEEE
  conference on computer vision and pattern recognition. pp. 2117--2125 (2017)

\bibitem{lin2014microsoft}
Lin, T.Y., Maire, M., Belongie, S., Hays, J., Perona, P., Ramanan, D.,
  Doll{\'a}r, P., Zitnick, C.L.: Microsoft coco: Common objects in context. In:
  European conference on computer vision. pp. 740--755. Springer (2014)

\bibitem{Lu_ECCV_2016}
Lu, C., Krishna, R., Bernstein, M., Fei-Fei, L.: Visual relationship detection
  with language priors. In: European conference on computer vision. pp.
  852--869. Springer (2016)

\bibitem{Lundberg_NIPS_2017}
Lundberg, S.M., Lee, S.I.: A unified approach to interpreting model
  predictions. In: Guyon, I., Luxburg, U.V., Bengio, S., Wallach, H., Fergus,
  R., Vishwanathan, S., Garnett, R. (eds.) Advances in Neural Information
  Processing Systems 30, pp. 4765--4774. Curran Associates, Inc. (2017)

\bibitem{Mao_CVPR_2016}
Mao, J., Huang, J., Toshev, A., Camburu, O., Yuille, A.L., Murphy, K.:
  Generation and comprehension of unambiguous object descriptions. In:
  Proceedings of the IEEE conference on computer vision and pattern
  recognition. pp. 11--20 (2016)

\bibitem{mikolov2013efficient}
Mikolov, T., Chen, K., Corrado, G., Dean, J.: Efficient estimation of word
  representations in vector space. arXiv preprint arXiv:1301.3781  (2013)

\bibitem{Oquab_CVPR_2015}
Oquab, M., Bottou, L., Laptev, I., Sivic, J.: Is object localization for
  free?-weakly-supervised learning with convolutional neural networks. In:
  Proceedings of the IEEE conference on computer vision and pattern
  recognition. pp. 685--694 (2015)

\bibitem{Peyre_ICCV_2019}
Peyre, J., Laptev, I., Schmid, C., Sivic, J.: Detecting unseen visual relations
  using analogies. In: Proceedings of the IEEE International Conference on
  Computer Vision. pp. 1981--1990 (2019)

\bibitem{Peyre_ICCV_2017}
Peyre, J., Sivic, J., Laptev, I., Schmid, C.: Weakly-supervised learning of
  visual relations. In: Proceedings of the IEEE International Conference on
  Computer Vision. pp. 5179--5188 (2017)

\bibitem{Plummer_ICCV_2017}
Plummer, B.A., Mallya, A., Cervantes, C.M., Hockenmaier, J., Lazebnik, S.:
  Phrase localization and visual relationship detection with comprehensive
  image-language cues. In: Proceedings of the IEEE International Conference on
  Computer Vision. pp. 1928--1937 (2017)

\bibitem{Pope_CVPR_2019}
Pope, P.E., Kolouri, S., Rostami, M., Martin, C.E., Hoffmann, H.:
  Explainability methods for graph convolutional neural networks. In:
  Proceedings of the IEEE Conference on Computer Vision and Pattern
  Recognition. pp. 10772--10781 (2019)

\bibitem{Prest_PAMI_2011}
Prest, A., Schmid, C., Ferrari, V.: Weakly supervised learning of interactions
  between humans and objects. IEEE Transactions on Pattern Analysis and Machine
  Intelligence  \textbf{34}(3),  601--614 (2011)

\bibitem{Qi_ECCV_2018}
Qi, S., Wang, W., Jia, B., Shen, J., Zhu, S.C.: Learning human-object
  interactions by graph parsing neural networks. In: Proceedings of the
  European Conference on Computer Vision (ECCV). pp. 401--417 (2018)

\bibitem{Qi_ICCV_2017}
Qi, X., Liao, R., Jia, J., Fidler, S., Urtasun, R.: 3d graph neural networks
  for rgbd semantic segmentation. In: Proceedings of the IEEE International
  Conference on Computer Vision. pp. 5199--5208 (2017)

\bibitem{ren2015faster}
Ren, S., He, K., Girshick, R., Sun, J.: Faster r-cnn: Towards real-time object
  detection with region proposal networks. In: Advances in neural information
  processing systems. pp. 91--99 (2015)

\bibitem{Ribeiro_SIGKDD_2016}
Ribeiro, M.T., Singh, S., Guestrin, C.: "why should {I} trust you?": Explaining
  the predictions of any classifier. In: Proceedings of the 22nd {ACM} {SIGKDD}
  International Conference on Knowledge Discovery and Data Mining, San
  Francisco, CA,USA, August 13-17, 2016. pp. 1135--1144 (2016)

\bibitem{russakovsky2015imagenet}
Russakovsky, O., Deng, J., Su, H., Krause, J., Satheesh, S., Ma, S., Huang, Z.,
  Karpathy, A., Khosla, A., Bernstein, M., et~al.: Imagenet large scale visual
  recognition challenge. International journal of computer vision
  \textbf{115}(3),  211--252 (2015)

\bibitem{Sadeghi_CVPR_2011}
Sadeghi, M., Farhadi, A.: Recognition using visual phrases. In: Proceedings of
  the 2011 IEEE Conference on Computer Vision and Pattern Recognition. pp.
  1745--1752 (2011)

\bibitem{Santoro_NIPS_2017}
Santoro, A., Raposo, D., Barrett, D.G., Malinowski, M., Pascanu, R., Battaglia,
  P., Lillicrap, T.: A simple neural network module for relational reasoning.
  In: Advances in neural information processing systems. pp. 4967--4976 (2017)

\bibitem{Selvaraju_ICCV_2017}
Selvaraju, R.R., Cogswell, M., Das, A., Vedantam, R., Parikh, D., Batra, D.:
  Grad-cam: Visual explanations from deep networks via gradient-based
  localization. In: Proceedings of the IEEE international conference on
  computer vision. pp. 618--626 (2017)

\bibitem{Tsubaki_BioInf_2019}
Tsubaki, M., Tomii, K., Sese, J.: Compound--protein interaction prediction with
  end-to-end learning of neural networks for graphs and sequences.
  Bioinformatics  \textbf{35}(2),  309--318 (2019)

\bibitem{wu2019detectron2}
Wu, Y., Kirillov, A., Massa, F., Lo, W.Y., Girshick, R.: Detectron2.
  \url{https://github.com/facebookresearch/detectron2} (2019)

\bibitem{xie2017aggregated}
Xie, S., Girshick, R., Doll{\'a}r, P., Tu, Z., He, K.: Aggregated residual
  transformations for deep neural networks. In: Proceedings of the IEEE
  conference on computer vision and pattern recognition. pp. 1492--1500 (2017)

\bibitem{Yan_AAAI_2018}
Yan, S., Xiong, Y., Lin, D.: Spatial temporal graph convolutional networks for
  skeleton-based action recognition. In: Thirty-second AAAI conference on
  artificial intelligence (2018)

\bibitem{Yao_CVPR_2010}
Yao, B., Fei-Fei, L.: Modeling mutual context of object and human pose in
  human-object interaction activities. In: 2010 IEEE Computer Society
  Conference on Computer Vision and Pattern Recognition. pp. 17--24. IEEE
  (2010)

\bibitem{Ying_NeurIPS_2019}
Ying, Z., Bourgeois, D., You, J., Zitnik, M., Leskovec, J.: Gnnexplainer:
  Generating explanations for graph neural networks. In: Advances in Neural
  Information Processing Systems. pp. 9240--9251 (2019)

\bibitem{Zhang_ICCV_2017}
Zhang, H., Kyaw, Z., Yu, J., Chang, S.F.: Ppr-fcn: Weakly supervised visual
  relation detection via parallel pairwise r-fcn. In: Proceedings of the IEEE
  International Conference on Computer Vision. pp. 4233--4241 (2017)

\bibitem{Zhang_IJCV_2018}
Zhang, J., Bargal, S.A., Lin, Z., Brandt, J., Shen, X., Sclaroff, S.: Top-down
  neural attention by excitation backprop. International Journal of Computer
  Vision  \textbf{126}(10),  1084--1102 (2018)

\bibitem{Zhou_CVPR_2016}
Zhou, B., Khosla, A., Lapedriza, A., Oliva, A., Torralba, A.: Learning deep
  features for discriminative localization. In: Proceedings of the IEEE
  conference on computer vision and pattern recognition. pp. 2921--2929 (2016)

\bibitem{Zhou_ICCV_2019}
Zhou, P., Chi, M.: Relation parsing neural network for human-object interaction
  detection. In: Proceedings of the IEEE International Conference on Computer
  Vision. pp. 843--851 (2019)

\bibitem{Zhou_CVPR_2018}
Zhou, Y., Zhu, Y., Ye, Q., Qiu, Q., Jiao, J.: Weakly supervised instance
  segmentation using class peak response. In: Proceedings of the IEEE
  Conference on Computer Vision and Pattern Recognition. pp. 3791--3800 (2018)

\end{thebibliography}

\ifappendix
\clearpage
\appendix

\section*{Supplementary material}

The following pages contain: 
\ref{app:datasets} more details about the three datasets used in this work, 
\ref{app:architecture-hyperparameters} more details about the architecture of the predicate classifier and hyperparameter optimization, 
\ref{app:additional-experiments} additional relationship detection experiments and ablation studies, and 
\ref{app:additional-results} additional qualitative results from the three datasets.

\section{Datasets}
\label{app:datasets}
\begin{table}[H]
    \centering
    \caption{Comparison of the datasets used in this work}
    \label{tab:app-dataset-comparison}
    \begin{tabular}{@{}rccccccc@{}}
    \toprule
                                    & \multicolumn{2}{c}{Number of images} & \multicolumn{3}{c}{Vocabulary size} & \multicolumn{2}{c}{Unique triplets} \\
                                    & Train             & Test             & Subject    & Predicate   & Object   & Train             & Test            \\ \midrule
    HICO-DET~\cite{Chao_WACV_2018} & 38118             & 9658             & 1          & 117         & 80       & 600               & 600             \\
    VRD~\cite{Lu_ECCV_2016}        & 4006              & 1001             & 100        & 70          & 100      & 6672              & 2741            \\
    UnRel~\cite{Peyre_ICCV_2017}   & -                 & 1071             & 100        & 70          & 100      & -                 & 76              \\ \bottomrule
    \end{tabular}
\end{table}

\subsection{HICO-DET}
\label{app:datasets-hico}
The Humans Interacting with Common Objects dataset~\cite{Chao_ICCV_2015}, in its detection version~\cite{Chao_WACV_2018}, is available at \url{http://www-personal.umich.edu/~ywchao/hico}.
The subject of the relationships is always a \textit{person}.
The object vocabulary is the same as MS-COCO~\cite{lin2014microsoft}.
Its predicates indicate human-object interactions, \eg \textit{carry}.
Some images from MS-COCO are also contained in HICO-DET, but the authors made sure that the test set of HICO-DET has no overlap with MS-COCO.
We warn future users to ignore the EXIF rotation tags present on some of the images, in fact all bounding boxes are annotated \wrt the non-rotated images.
See \tabref{tab:app-dataset-comparison} for a comparison of dataset and vocabulary size.

\noindent We use the pre-trained object detector made available through the \texttt{detectron2} implementation~\cite{wu2019detectron2} of Faster~R-CNN~\cite{ren2015faster}.
Since the object detector is an important part of visual relationship detection pipelines, we report object detection metrics obtained for this dataset in \tabref{tab:app-object-detection-metrics}.

\begin{figure}[H]
    \centering%
    \captionsetup[subfigure]{justification=centering}%
    \hfill%
    \begin{subfigure}[b]{.33\textwidth}%
        \centering%
        \includegraphics[width=\linewidth]{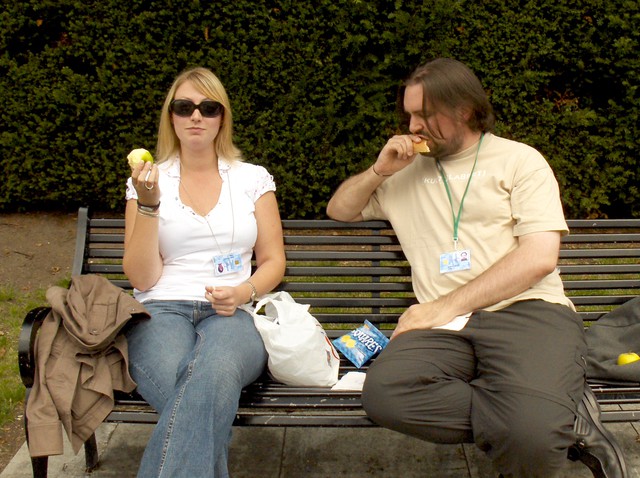}
        \vspace{-.6cm}\caption*{\tiny person eat sandwich\\person sit on bench}%
    \end{subfigure}%
    \hfill%
    \begin{subfigure}[b]{.379\textwidth}%
        \centering%
        \includegraphics[width=\linewidth]{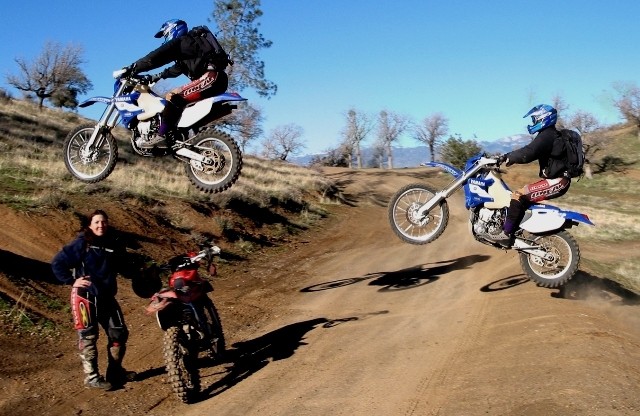}
        \vspace{-.6cm}\caption*{\tiny person drive motorcycle\\person hold motorcycle}%
    \end{subfigure}%
    \hspace*{\fill}%
    \caption{\footnotesize Ground-truth triplet annotations from the HICO-DET dataset}%
    \label{fig:app-example-hico}%
\end{figure}

\subsection{Visual Relationship Detection dataset}
\label{app:datasets-vrd}
The Visual Relationship Detection Dataset (VRD)~\cite{Lu_ECCV_2016} is available at \url{https://cs.stanford.edu/people/ranjaykrishna/vrd}.
Its images and annotations correspond to those in the Scene Graph dataset \cite{Johnson_CVPR_2015}, but the vocabularies of objects and predicates have been carefully curated, \eg \figref{fig:app-example-vrd}
We warn future users to ignore the EXIF rotation tags present on some of the images, in fact all bounding boxes are annotated \wrt the non-rotated images.
Also, we note that for some images the annotation file contains 0 objects and 0 relationships.
See \tabref{tab:app-dataset-comparison} for a comparison of dataset and vocabulary size.

Since no pre-trained model is publicly available for this dataset, we fine-tune an object detector based on \texttt{detectron2}~\cite{wu2019detectron2}.
Object detection metrics are reported in \tabref{tab:app-object-detection-metrics} for future reference.

\begin{figure}[H]
    \centering%
    \captionsetup[subfigure]{justification=centering}%
    \hfill%
    \begin{subfigure}[b]{.325\textwidth}%
        \centering%
        \includegraphics[width=\linewidth]{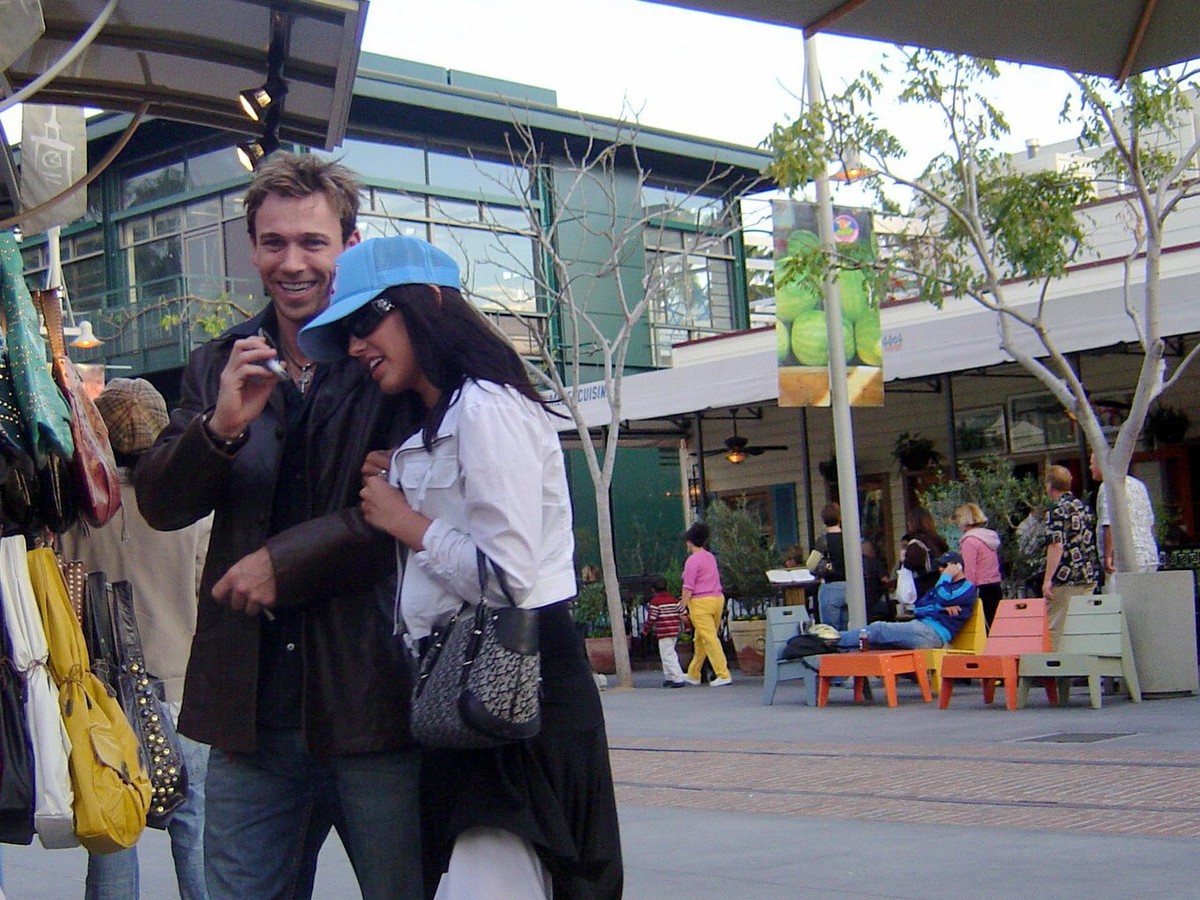}
        \vspace{-.6cm}\caption*{\tiny person has cellphone\\tree in the front of building}%
    \end{subfigure}%
    \hfill%
    \begin{subfigure}[b]{.365\textwidth}%
        \centering%
        \includegraphics[width=\linewidth]{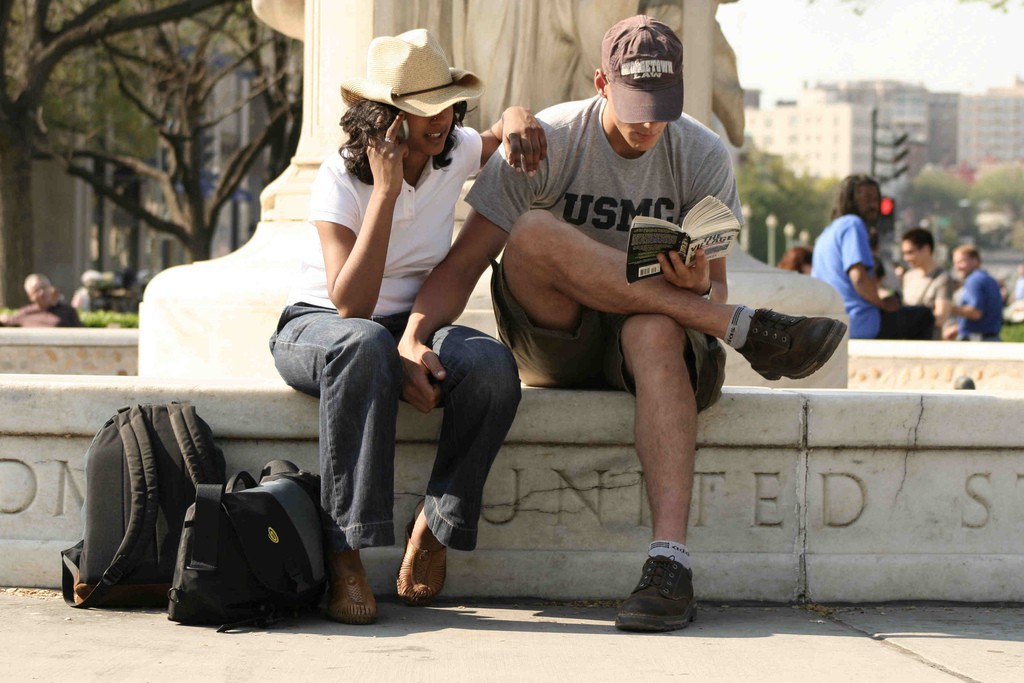}
        \vspace{-.6cm}\caption*{\tiny person read book\\backpack next to person}%
    \end{subfigure}%
    \hspace*{\fill}%
    \caption{\footnotesize Ground-truth triplet annotations from the VRD dataset}%
    \label{fig:app-example-vrd}%
\end{figure}

\subsection{Unusual Relationships dataset}
\label{app:datasets-unrel}
The Unusual Relationships dataset (UnRel)~\cite{Peyre_ICCV_2017} is available at \url{https://www.di.ens.fr/willow/research/unrel}.
It is meant as an evaluation-only dataset for rare and unusual relationships, \eg \figref{fig:app-example-unrel}.
See \tabref{tab:app-dataset-comparison} for a comparison of dataset and vocabulary size.

Since it shares the same object and predicate vocabulary of VRD, we use the same object detector, of which we report object detection metrics in \tabref{tab:app-object-detection-metrics}

\begin{figure}[H]
    \centering%
    \hfill%
    \begin{subfigure}[b]{.405\textwidth}%
        \centering%
        \includegraphics[width=\linewidth]{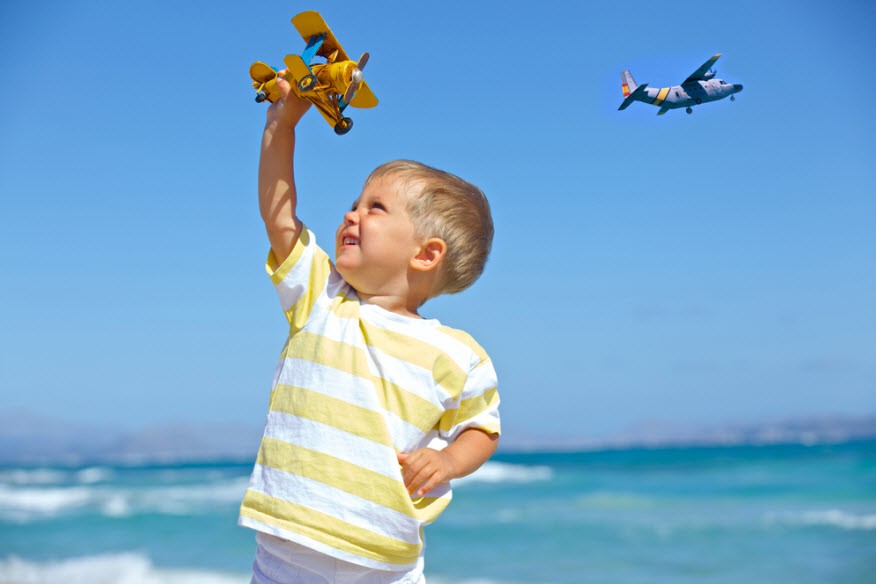}
        \vspace{-.6cm}\caption*{\tiny person hold plane}%
    \end{subfigure}%
    \hfill%
    \begin{subfigure}[b]{.27\textwidth}%
        \centering%
        \includegraphics[width=\linewidth]{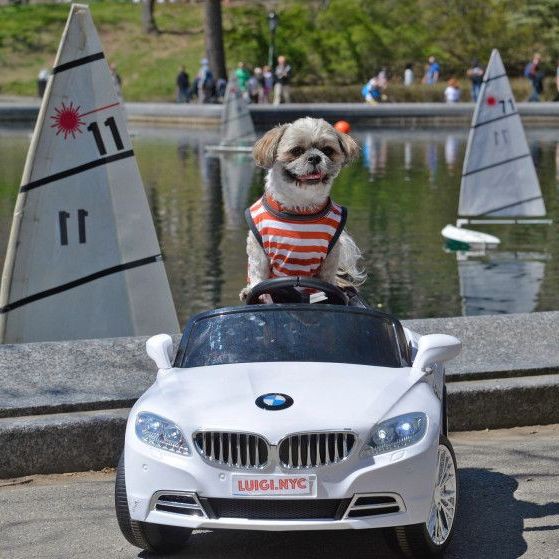}
        \vspace{-.6cm}\caption*{\tiny dog drive car}%
    \end{subfigure}%
    \hspace*{\fill}%
    \caption{\footnotesize Ground-truth triplet annotations the UnRel dataset}%
    \label{fig:app-example-unrel}%
\end{figure}

\begin{table}[th]
    \setlength{\tabcolsep}{.6em}
    \centering
    \caption{Object detection metrics for the datasets used in this work}
    \label{tab:app-object-detection-metrics}
    \begin{tabular}{@{}lcccccc@{}}
    \toprule
                                    & \multicolumn{6}{c}{Mean Average Precision}                   \\
                                    & IoU@[0.5:0.95] & IoU@0.5 & IoU@0.75 & small & medium & large \\ \midrule
    HICO-DET~\cite{Chao_WACV_2018} & 20.2           & 34.1    & 20.8     & 2.3   & 11.5   & 29.7  \\
    VRD~\cite{Lu_ECCV_2016}        & 21.2           & 35.3    & 22.6     & 4.9   & 14.3   & 25.0  \\
    UnRel~\cite{Peyre_ICCV_2017}   & 21.0           & 35.3    & 22.6     & 4.9   & 14.3   & 25.0  \\ \bottomrule
    \end{tabular}%
    \vspace{1.5em}
    \begin{tabular}{@{}lcccccc@{}}
    \toprule
                                    & \multicolumn{6}{c}{Mean Average Recall}           \\
                                    & top-1 & top-10 & top-100 & small & medium & large \\ \midrule
    HICO-DET~\cite{Chao_WACV_2018} & 30.3  & 39.3   & 40.2    & 11.6  & 29.2   & 48.6  \\
    VRD~\cite{Lu_ECCV_2016}        & 34.0  & 45.0   & 45.1    & 14.9  & 33.2   & 48.3  \\
    UnRel~\cite{Peyre_ICCV_2017}   & 34.0  & 45.0   & 45.1    & 14.9  & 33.2   & 48.3  \\ \bottomrule
    \end{tabular}
\end{table}

\section{Architecture and hyperparameters}
\label{app:architecture-hyperparameters}

\subsection{Introduction to GNNs}
\label{app:intro-gnn}
In our work, an image is first represented as a fully-connected graph of objects and then processed through a graph neural network to predict predicates. 
Specifically, we use a message-passing implementation of graph convolution.
At the input, each node $i$ is associated to a feature vector $\vv_i$. 
Similarly, each edge $i \rightarrow j$ is associated to a feature vector $\ve_{i,j}$. 
A global bias term $\vu$ can be used to represent information that is not localized to any specific node/edge of the graph. 
With this graph representation, one layer of message passing performs the following updates.
\begin{enumerate}
\item For every edge $i \rightarrow j$, the edge vector is updated using a function $f^e$ that takes as input the adjacent nodes $\vv_i$ and $\vv_j$, the edge itself $\ve_{i,j}$ and the global attribute $\vu$:
      $$\ve'_{i,j} = f^e\left(\vv_i, \vv_j, \ve_{i,j}, \vu \right)$$
\item For every node $i$, features from incident edges $\set{\ve'_{j,i}}$ are aggregated using a pooling function $\text{agg}^{e \rightarrow v}$:
      $$\bm{\bar{e}}'_i = \text{agg}^{e \rightarrow v} \set{\ve'_{j,i}} $$
\item For every node $i$, the node feature vector is updated using a function $f^v$ that takes as input the aggregated incident edges $\bm{\bar{e}}'_i$, the node itself $\vv_i$ and the global attribute $\vu$:
      $$\vv'_i = f^v\left(\bm{\bar{e}}'_i, \vv_i, \vu\right)$$
\item All edges are aggregated using a pooling function $\text{agg}^{e \rightarrow u}$:
      $$\bm{\bar{e}}' = \text{agg}^{e \rightarrow u} \set{\ve'_{i,j}} $$
\item All nodes are aggregated using a pooling function $\text{agg}^{v \rightarrow u}$:
      $$\bm{\bar{v}}' = \text{agg}^{v \rightarrow u} \set{\vv'_i} $$
\item The global feature vector is updated using a function $f^u$ of the aggregated edges $\bm{\bar{e}}'$, of the aggregated nodes $\bm{\bar{v}}'$ and of the global attribute $\vu$:
      $$\vu' = f^u\left(\bm{\bar{e}}', \bm{\bar{v}}', \vu\right)$$
\end{enumerate}

These convolutional layers can be stacked to increase the receptive fields of a node. However, in this work, we used a single layer to focus on pairwise relationships. Furthermore, we did not use a global attribute $\vu$, which could encode for example context and background.

\subsection{Predicate classifier}
\label{app:predicate-classifier-hyperparameters}
For the predicate classifier we optimize the hyperparameters reported in \tabref{tab:app-hyperparams-pred-class}.
Rather than performing a grid-search over the whole space, we perform a "guided" search:
we iteratively perform parallel runs and only keep the best-performing combinations of parameters.
This process of trial and elimination allows us to quickly prune unpromising regions of the search space.

\begin{table}[H]
    \centering
    \caption{Hyperparameter space of the predicate classifier}
    \label{tab:app-hyperparams-pred-class}
    \begin{tabular}{@{}lll@{}}
    \toprule
    Parameter                       & Choices                             & Final value \\ \midrule
    \textbf{Optimizer}              &                                     &             \\
    \quad Learning rate                   & $10^{-2}, 10^{-3}, 10^{-4}\qquad$  & $10^{-3}$   \\
    \quad Weight decay                    & $10^{-3}, 10^{-5}, 0$               & $10^{-5}$   \\
    \quad Max epochs                      & 35                                  & 18          \\
    \textbf{Model}                  &                                     &             \\
    \quad Linear layers                   & 1, 2                                & 1           \\
    \quad Linear features                 & 256, 512, 1024                      & 1024        \\
    \quad Convolutional layers            & 1, 2                                & 2           \\
    \quad Convolutional kernels$\qquad$   & 256, 512                            & 256         \\
    \quad Pooling function                & add, max, mean                      & max         \\
    \quad Bias in $f_p$                   & yes, no                             & yes         \\ \bottomrule
    \end{tabular}
\end{table}

The best set of hyperparameters is chosen to maximize \texttt{recall@5} over a held-out validation set (15\% of training data).
The train/val split is made at random for every training run. Random seeds are fixed at the beginning of each run and recorded for reproducibility.
Note that \texttt{recall@5} refers to the image-level predicate predictions, and relationship detection metrics are not involved in the optimization of the predicate classifier.

On the test set of HICO-DET, relative to predicate classification only, these parameters achieve a mAP of 0.44, \texttt{recall@5} of 0.90 and \texttt{recall@10} of 0.96.

\subsection{ResNeXt baseline and Grad-CAM}
\label{app:resnext-baseline}
We finetune a ResNeXt-50 \cite{xie2017aggregated} for predicate classification on the Visual Relationship Detection dataset.
All parameters are initialized from an ImageNet \cite{russakovsky2015imagenet} pretraining, except the final classification layer that is adapted to output 70-dimensional vector of predicate predictions and is initialized from a Normal distribution.
Given an input image $\Image \in [0,1]^{3\times H\times W}$, the convolutional architecture can be summarized as:
\begin{align}
    \vh &= \textsc{ResNeXt}(\Image) \in \mathbb{R}^{2048 \times \tilde{H} \times \tilde{W}} &\text{\footnotesize backbone}\\
    \evz_c &= \frac{1}{\tilde{H}\tilde{W}} \sum_{i=1}^{\tilde{H}} \sum_{j=1}^{\tilde{W}} \evh_{c,i,j} \quad \forall c = 1,\dots,2048 &\text{\footnotesize global average pooling} \\
    \vy &= \text{softmax}(\mW \vz + \vb) \in [0, 1]^K & \text{\footnotesize classification}
\end{align}
where $\tilde{H}$ and $\tilde{W}$ represent the height and width of the feature volume extracted by the backbone before global average pooling.

We use Adam optimizer \cite{kingma2014adam} to minimize the same loss of the GNN-based predicate classifier described in the main text.
The learning rate is set to $10^{-3}$ for the classification layer and to $10^{-4}$ for the rest of the network.

We optimize only the number of epochs and whether the final layer should include a bias term or not.
Based on performances on the validation set, the best hyperparameters are training for 6 epochs and including the bias.
The final CNN-based model achieves similar \texttt{recall@5} as the GNN-based classifier on the test set for predicate classification.

\noindent Grad-CAM heatmaps as in \figref{fig:method-gradcam} are produced by computing:
\begin{align}
    \alpha^k_c &= \frac{1}{\tilde{H}\tilde{W}} \sum_{i=1}^{\tilde{H}} \sum_{j=1}^{\tilde{W}} \frac{\partial \evy_k}{\partial \evh_{c,i,j}} \quad \forall c = 1,\dots,2048\\
    \evs_{i,j} &=\ReLU\left( \sum_{c=1}^{2048} \alpha^k_c \evh_{c,i,j} \right) \forall i = 1,\dots,\tilde{H}; j = 1,\dots,\tilde{W}.
\end{align}
Then the 2D vector $\vs$ is upsampled to the $H\times W$ size of the input image, and its values are normalized to the range $[0, 1]$.

\subsection{Training and inference}
The graph neural network described in \secref{sec:method-predicate-classification} is trained to classify the predicates present in an image from image-level annotations.

\begin{algorithm}[H]
\SetAlgoLined
\DontPrintSemicolon
\KwIn{%
Pretrained object detector (\texttt{detectron2}),\newline
Dataset of images with image-level predicate annotations.
}
\Repeat{convergence}{
    Extract objects from image $\Image$\;
    Build a fully-connected image graph $\Graph$ using features from eq. \ref{eq:spatial-features}, \ref{eq:visual-features}\;
    Apply the predicate classifier to $\Graph$\;
    Compute the predicate classification loss $\Loss$ (\eqref{eq:pred-class-loss})\;
    Minimize $\Loss$ using Adam optimizer\;
}
\KwOut{Trained predicate classifier}
\caption{Training Algorithm}
\end{algorithm}
\vspace{1em}
\noindent Once trained, the predicate classifier can be used for relationship detection. 
Specifically, each $\pred$ prediction is attributed to pairs of objects in the input by means of explanation, thus retrieving the full $\langle\subj,\pred,\obj\rangle$ triplet.

\begin{algorithm}[H]
\SetAlgoLined
\DontPrintSemicolon
\KwIn{%
Pretrained object detector (\texttt{detectron2}),\newline
Trained predicate classifier,\newline
Image of interest $\Image$.
}
\uIf{Predicate Detection}{
    Extract ground-truth objects from image $\Image$\;
}
\ElseIf{Phrase Detection $\lor$ Relationship Detection}{
    Detect objects in $\Image$ using the object detector\;
}
Build a fully-connected scene graph $\Graph$ using features from eq. \ref{eq:spatial-features}, \ref{eq:visual-features}\;
Apply the predicate classifier to $\Graph$\;
Visual relations $\mathcal{R} \gets \emptyset$\;
\For{$\pred \in \set{N \text{top-scoring predicates}}$} {
    \tcc{Predicate predictions are explained in terms of relevant pairs of objects in the image graph $\Graph$}
    Compute node and edge relevances using eq. \ref{eq:relevance-nodes}, \ref{eq:relevance-edges}\;
    Score each $\langle\subj,\obj\rangle$ pair using \eqref{eq:visual-relation-explanation-approx}\;
    Multiply the score by the object detection scores of $\subj$ and $\obj$\;
    Multiply the score by the classification score of $\pred$\;
    Multiply the score by the relationship prior (\eqref{eq:prior-over-relations})\;
    Store high-scoring triplets $\langle\subj,\pred,\obj\rangle$ in $\mathcal{R}$\;
}
\KwOut{$K$ top-scoring visual relations from $\mathcal{R}$}
\caption{Explanation-based Relationship Detection Algorithm}
\end{algorithm}

\clearpage
\section{Additional experiments}
\label{app:additional-experiments}

\subsection{Pooling function}
As explained in \appref{app:architecture-hyperparameters}, the pooling function for \eqref{eq:aggr-readout-fn} is selected according to predicate classification performances (\figref{fig:app-pooling-fn-predicate-classification}) on a 15\% split of the training set.
\Figref{fig:app-pooling-fn-predicate-classification} shows \texttt{recall@5} for \textit{sum}, \textit{max}, and \textit{mean} pooling over 10 runs on the VRD dataset.
Due to higher recall on the validation set, \textit{max} pooling is selected and used for all results reported in the main text.
We notice, however, that \textit{mean} pooling performs closely to \textit{max}.

\begin{figure}[H]
    \centering
    \vspace{-1.5em}
    \includegraphics[width=.6\linewidth]{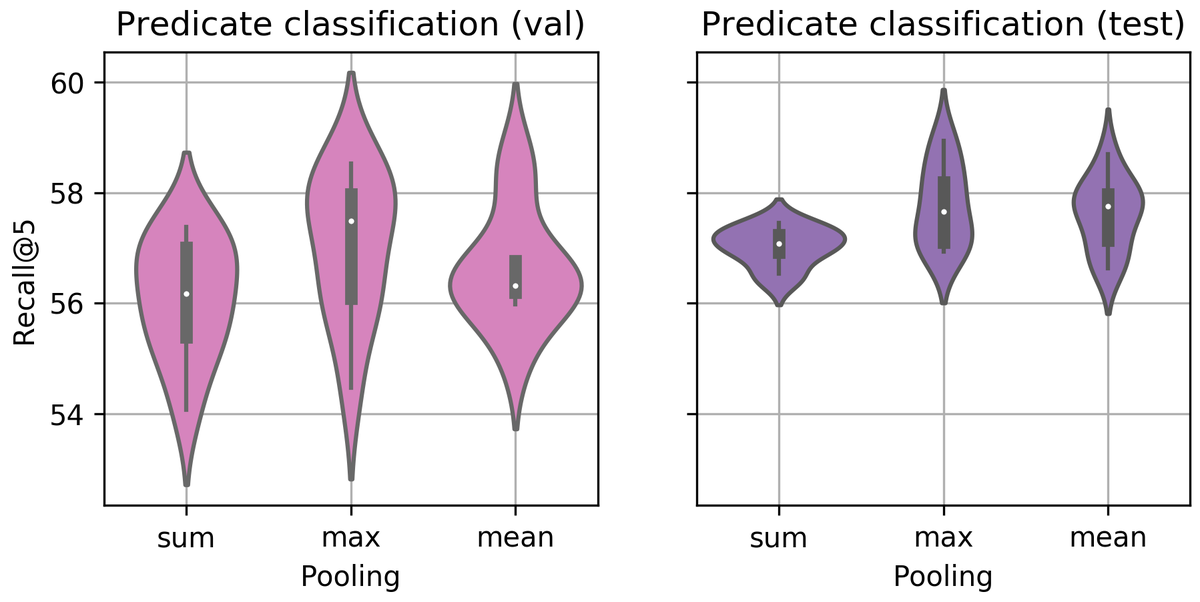}
    \vspace{-1em}\caption{\texttt{Recall@5} for predicate classification on VRD using different pooling functions. Validation set (15\% of training) on the left, and test set on the right}
    \label{fig:app-pooling-fn-predicate-classification}\vspace{-1em}
\end{figure}

To further test the role of pooling, we evaluated relationship detection metrics for several predicate classifiers trained using \textit{sum}, \textit{max}, and \textit{mean} pooling.
\Figref{fig:app-pooling-fn-relationship-detection} shows that \textit{mean} pooling outperforms the other two, despite performing slightly worse for predicate classification.

\begin{figure}[H]
    \centering
    \vspace{-1em}
    \includegraphics[width=\linewidth]{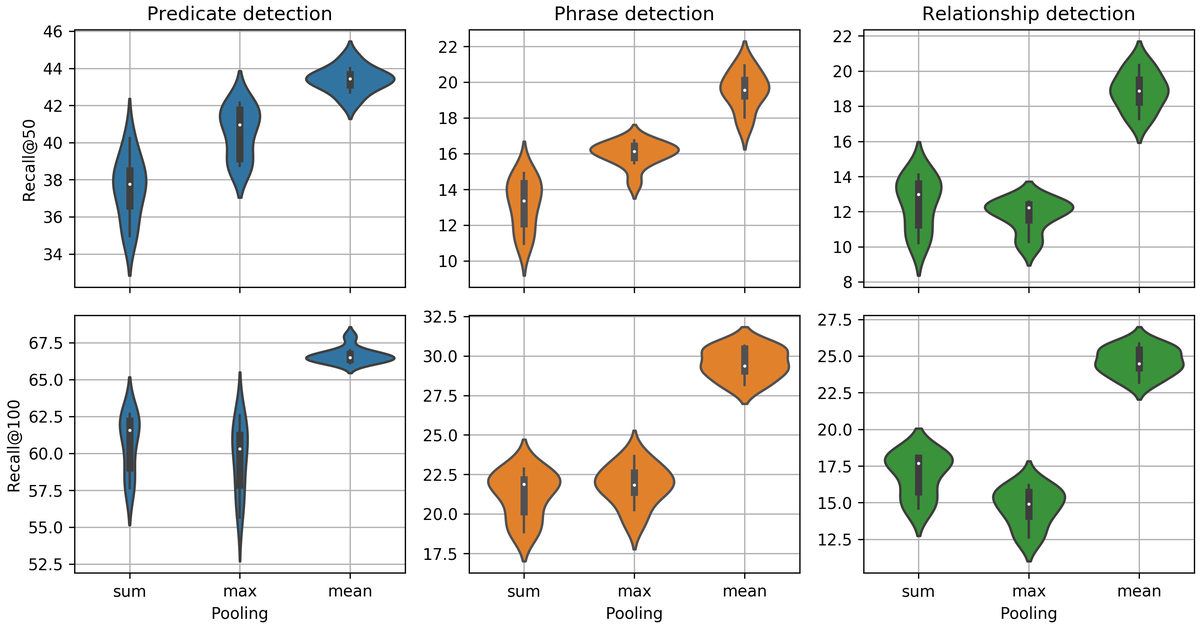}
    \caption{\texttt{Recall@50} and \texttt{@100} for relationship detection on VRD using different pooling functions. \textit{mean} pooling outperforms the other two, despite performing slightly worse for predicate classification}
    \label{fig:app-pooling-fn-relationship-detection}
\end{figure}

\subsection{Number of explained predicates}
Given an image, the GNN classifier outputs a distribution of binary probabilities over the predicates contained in the image.
To recover $\langle\subj,\pred,\obj\rangle$ triplets, we consider the top $N$ predicates and \textit{explain} them one at the time \wrt the input image graph.
Therefore, the choice of $N$ influences the diversity of predicates contained in the detected relationships, \eg if we only explained the top scoring predicate we could still recover many triplets but they would all share the same predicate.

For the main results, we set $N=10$, assuming that in natural images the chance of having more than ten different predicates depicted in the same picture would be rather low.
To further prove this point, in \figref{fig:app-top-n-predicates} we plot \texttt{recall@50} and \texttt{recall@100} for various choices of $N$ on the VRD dataset.
Notably, considering very few predicates in the explanation phase, gives poor results on all three relationship detection scenarios.
However, increasing $N$ to consider more predicate categories yields diminishing returns after $N=20$.
\begin{figure}[H]
    \centering
    \includegraphics[width=\linewidth]{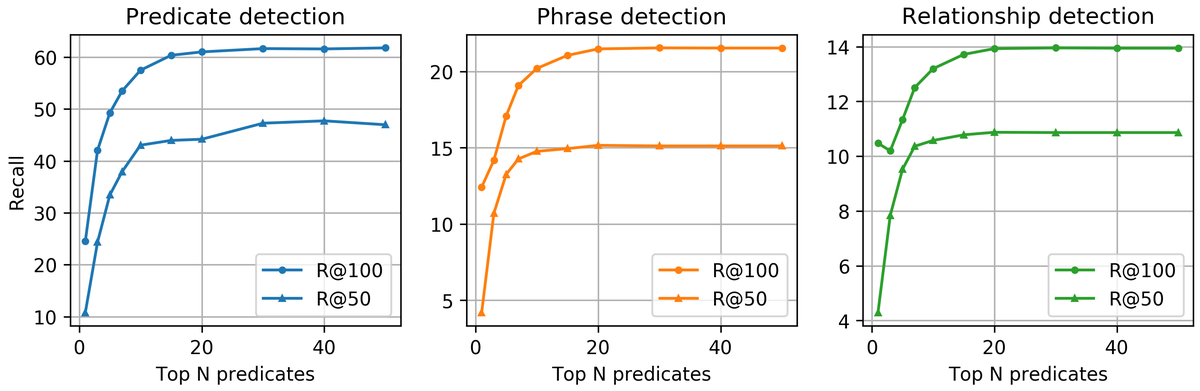}
    \caption{Recall at 50 (R@50) and at 100 (R@100) on the VRD dataset as the number $N$ of predicates considered for explanation increases from 1 to 50. Diminishing returns are observed, with an elbow at approximately $N=10$}
    \label{fig:app-top-n-predicates}
\end{figure}

\subsection{Relationship prior}
\label{app:relationship-prior}
As explained in \secref{sec:method-prior-over-relations}, a weakly-supervised method trained only on predicate labels is not able to learn the directionality of the relations, \eg it could not distinguish \textit{car on street} from \textit{street on car}.
Therefore, we introduced a simple relationship prior based on the frequency of relationships in a small subset of training data. Specifically, we compute:
\begin{equation*}
\text{freq}(c_i, c_j | k) = \frac{|\set{\langle c_\subj, b_\subj, k, c_\obj, b_\subj \rangle | c_\subj=c_i, c_\obj=c_j, k_\pred=k}|}{|\set{\langle c_\subj, b_\subj, k, c_\obj, b_\subj \rangle | k_\pred=k}|}
\end{equation*}

In the main experiments, we use a 15\% split of the training set to compute this prior, assuming that it would be enough to disambiguate most cases.
In \figref{fig:app-frequencies}, we show how \texttt{recall@50} and \texttt{recall@100} on the VRD dataset change according to the percentage of training triplets used to compute the relationship prior.
For each percentage value, we plot the mean recall over 5 random subsets and shade the area corresponding to two standard deviations.
We observe that all percentages obtain approximately the same recall, except for 0\% that corresponds to a uniform prior.
Notably, the randomness introduced when choosing a subset of the given percentage of training data has little effect on the result.
\begin{figure}[H]
    \centering
    \includegraphics[width=\linewidth]{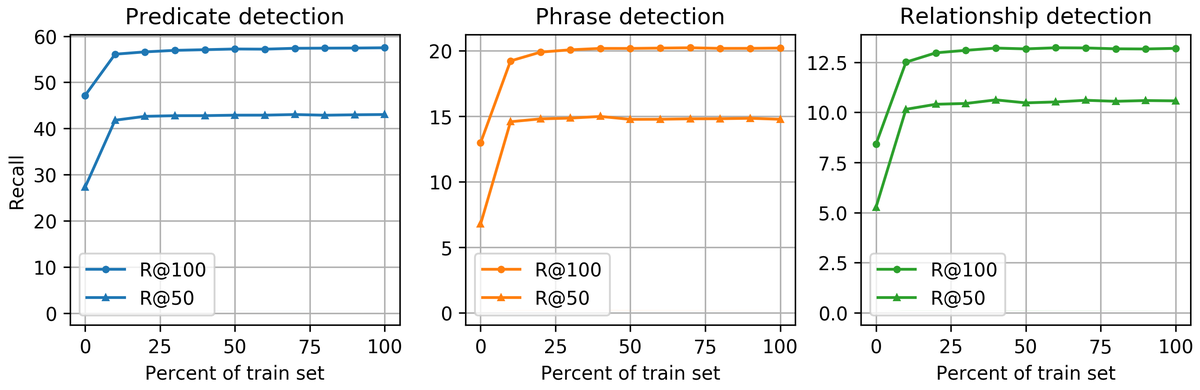}
    \caption{Recall at 50 (R@50) and at 100 (R@100) on the VRD dataset as the percentage of training data used to compute the relationship prior increases. At each percentage, we run 5 evaluations and plot mean and two standard deviations. Each evaluation uses a different random subset to compute the prior. All percentages obtain approximately the same recall, except for 0\% that corresponds to a uniform prior}
    \label{fig:app-frequencies}
\end{figure}

\clearpage
\section{Additional results}
\label{app:additional-results}

In this section we report additional qualitative results to evaluate the relationship detection pipeline.
We include examples of: correct relationship detections, correct detections missing from the ground truth, incorrect detections due to object misclassification, and incorrect detection due to subject-object inversion, wrong choice of pair, or wrong predicate.
All images in figures \ref{fig:app-additional-results-hico}, \ref{fig:app-additional-results-vrd} and \ref{fig:app-additional-results-unrel} are chosen at random from the test sets of each dataset.
Then, representative examples are chosen from the top 10 detections of each image (top 25 for UnRel).

\begin{figure}[ht]
    \centering
    \captionsetup[subfigure]{justification=centering}
    \begin{subfigure}[t]{.23\linewidth}
        \centering
        \includegraphics[width=\linewidth]{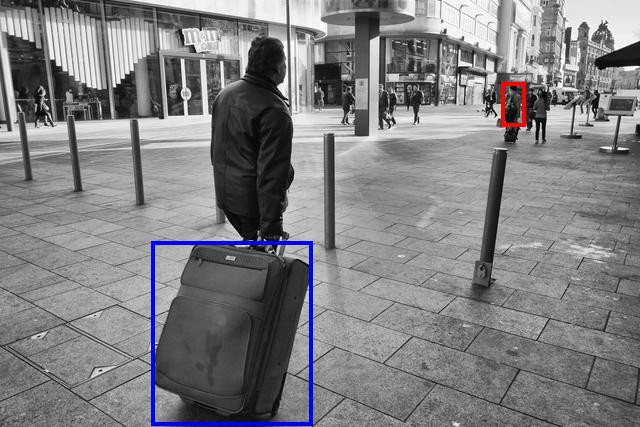}
        \vspace{-.6cm}\caption*{\tiny person hold suitcase}
    \end{subfigure}\hfill
    \begin{subfigure}[t]{.23\linewidth}
        \centering
        \includegraphics[width=\linewidth]{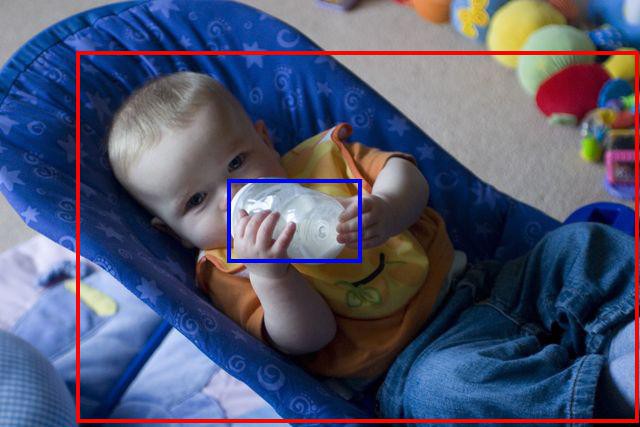}
        \vspace{-.6cm}\caption*{\tiny person sip bottle}
    \end{subfigure}\hfill
    \begin{subfigure}[t]{.23\linewidth}
        \centering
        \includegraphics[width=\linewidth]{figures/hico/empty.png}
        \vspace{-.6cm}\caption*{\tiny (not possible with ground-truth objects)}
    \end{subfigure}\hfill
    \begin{subfigure}[t]{.23\linewidth}
        \centering
        \includegraphics[width=\linewidth]{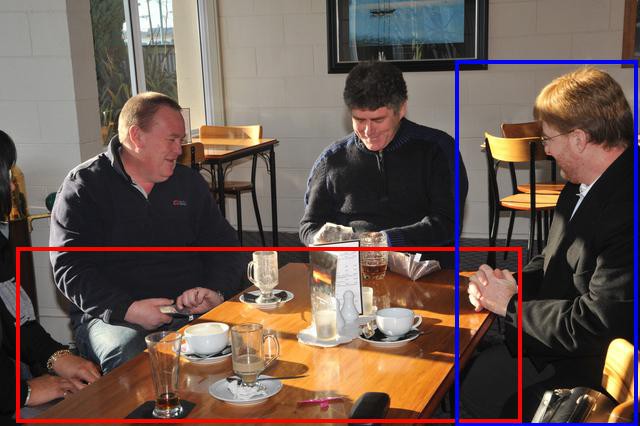}
        \vspace{-.6cm}\caption*{\tiny dining table eat at pers. \\(subj-obj inversion)}
    \end{subfigure}\\
    \begin{subfigure}[t]{.23\linewidth}
        \centering
        \includegraphics[width=\linewidth]{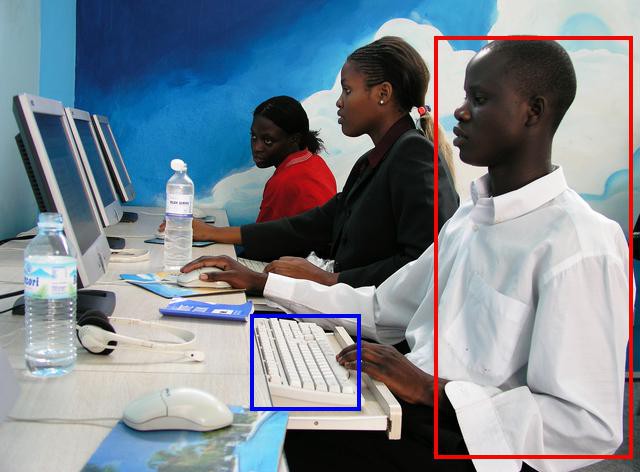}
        \vspace{-.6cm}\caption*{\tiny p. type on keyboard}
    \end{subfigure}\hfill
    \begin{subfigure}[t]{.23\linewidth}
        \centering
        \includegraphics[width=\linewidth]{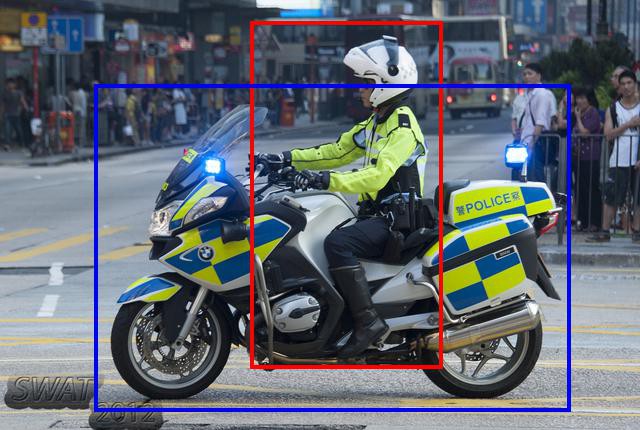}
        \vspace{-.6cm}\caption*{\tiny p. straddle motorcycle}
    \end{subfigure}\hfill
    \begin{subfigure}[t]{.23\linewidth}
        \centering
        \includegraphics[width=\linewidth]{figures/hico/empty.png}
        \vspace{-.6cm}\caption*{\tiny (not possible with ground-truth objects)}
    \end{subfigure}\hfill
    \begin{subfigure}[t]{.23\linewidth}
        \centering
        \includegraphics[width=\linewidth]{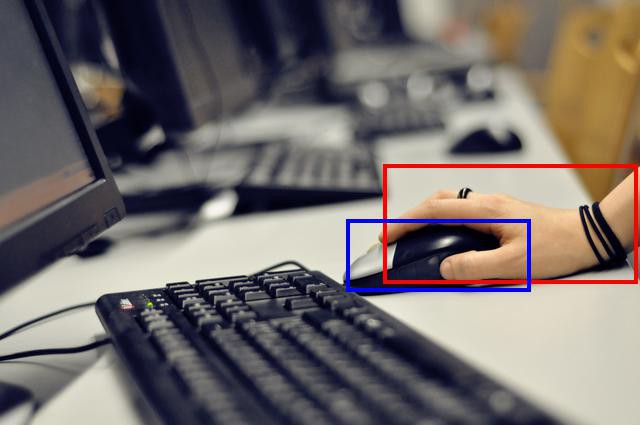}
        \vspace{-.6cm}\caption*{\tiny person check mouse \\(wrong predicate)}
    \end{subfigure}\\
    \begin{subfigure}[t]{.23\linewidth}
        \centering
        \includegraphics[width=\linewidth]{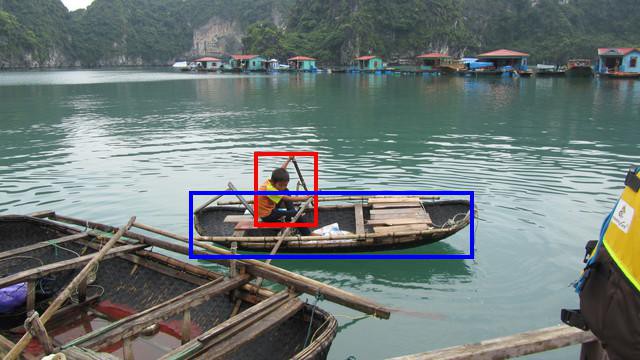}
        \vspace{-.6cm}\caption*{\tiny person row boat}
    \end{subfigure}\hfill
    \begin{subfigure}[t]{.23\linewidth}
        \centering
        \includegraphics[width=\linewidth]{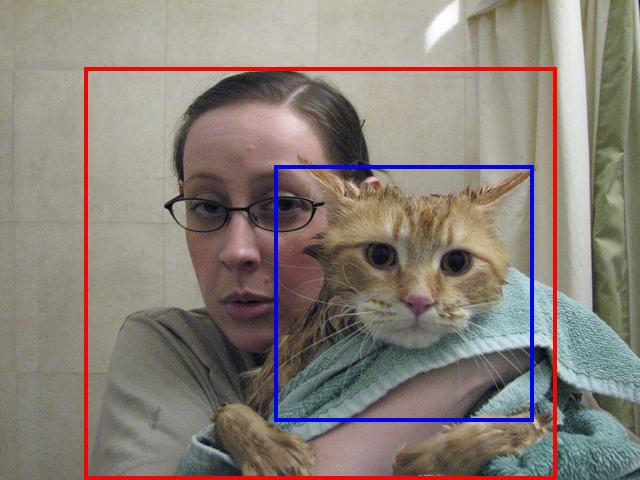}
        \vspace{-.6cm}\caption*{\tiny p. hug cat}
    \end{subfigure}\hfill
    \begin{subfigure}[t]{.23\linewidth}
        \centering
        \includegraphics[width=\linewidth]{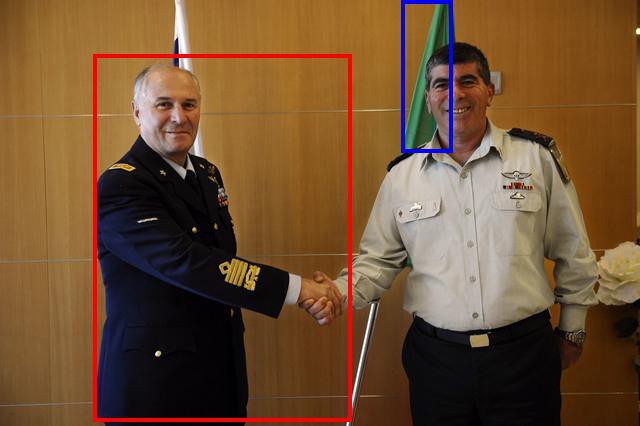}
        \vspace{-.6cm}\caption*{\tiny p. wield umbrella\\(correct object: flag)}
    \end{subfigure}\hfill
    \begin{subfigure}[t]{.23\linewidth}
        \centering
        \includegraphics[width=\linewidth]{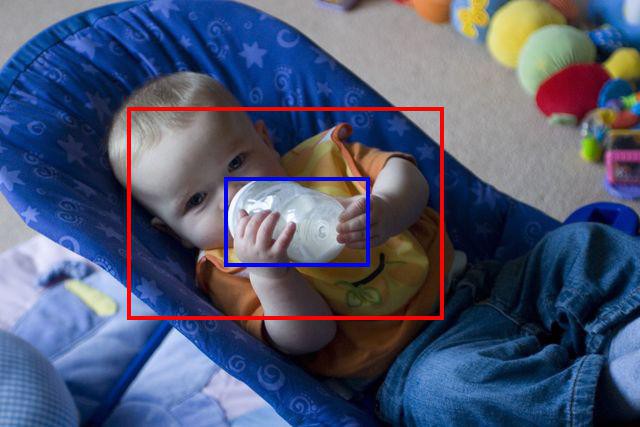}
        \vspace{-.6cm}\caption*{\tiny p. pet bottle \\(wrong predicate)}
    \end{subfigure}\\
    \begin{subfigure}[t]{.23\linewidth}
        \centering
        \includegraphics[width=\linewidth]{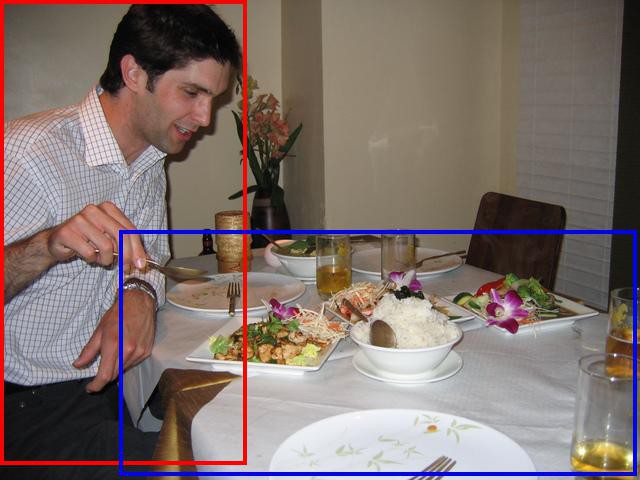}
        \vspace{-.6cm}\caption*{\tiny p. eat at dining table}
    \end{subfigure}\hfill
    \begin{subfigure}[t]{.23\linewidth}
        \centering
        \includegraphics[width=\linewidth]{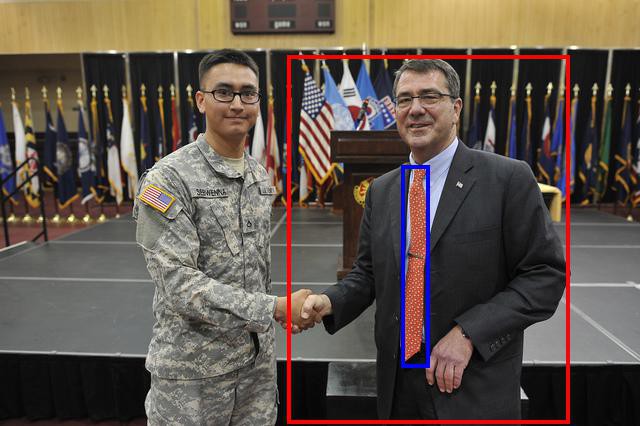}
        \vspace{-.6cm}\caption*{\tiny person wear tie}
    \end{subfigure}\hfill
    \begin{subfigure}[t]{.23\linewidth}
        \centering
        \includegraphics[width=\linewidth]{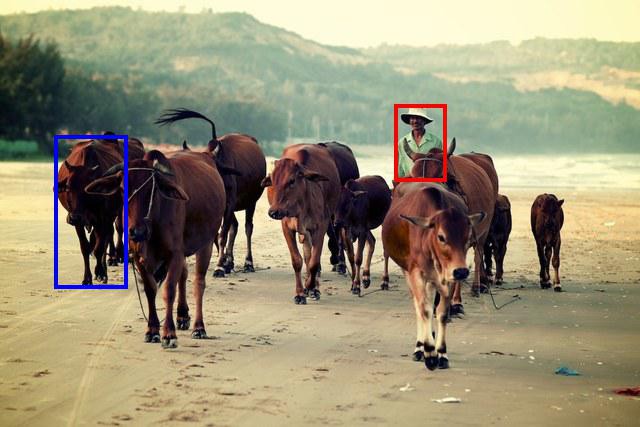}
        \vspace{-.6cm}\caption*{\tiny p. no interaction horse \\(correct object: cow)}
    \end{subfigure}\hfill
    \begin{subfigure}[t]{.23\linewidth}
        \centering
        \includegraphics[width=\linewidth]{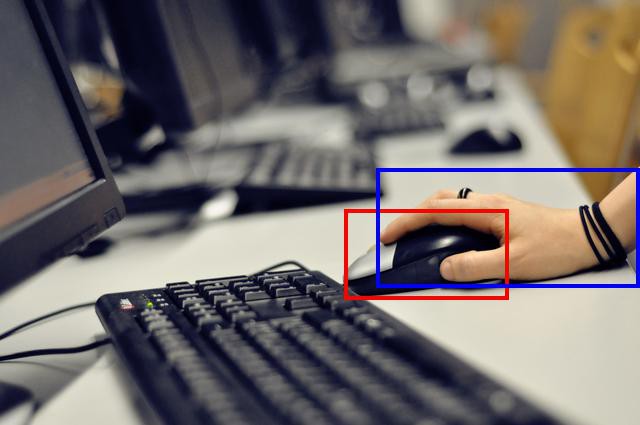}
        \vspace{-.6cm}\caption*{\tiny mouse control person \\(subj-obj inversion)}
    \end{subfigure}    
    \caption{\small \textbf{Additional detections on HICO-DET.} Top two rows use ground-truth objects, bottom two rows use Faster R-CNN objects. Subjects are framed in red, objects in blue. Left to right: correct relationship detection, correct but missing ground-truth, incorrect due to object misdetection, incorrect detection. Images are chosen at random from the test set, all depicted triplets are selected from the top 10 detections}
    \label{fig:app-additional-results-hico}
\end{figure}

\begin{figure}[ht]
    \centering
    \captionsetup[subfigure]{justification=centering}
    \begin{subfigure}[t]{.23\linewidth}
        \centering
        \includegraphics[width=\linewidth]{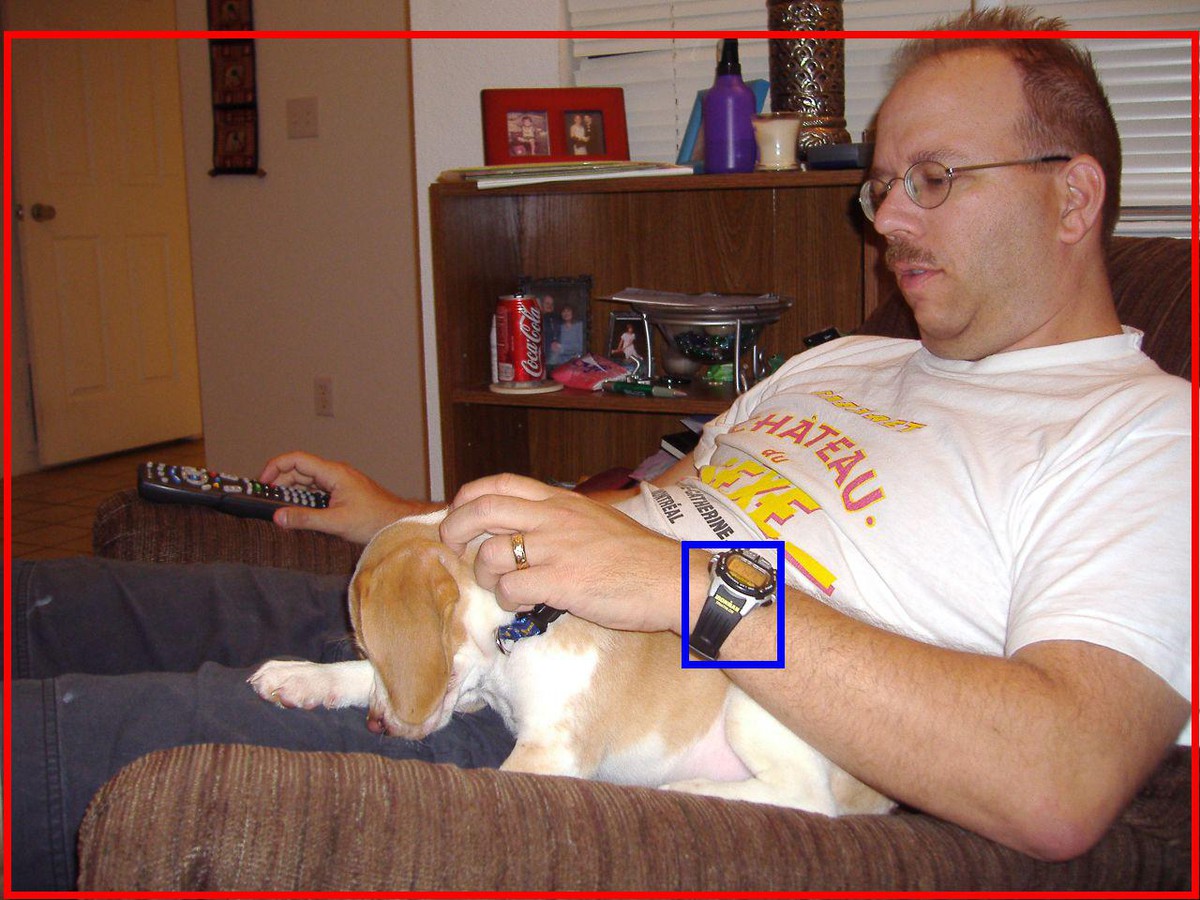}
        \vspace{-.6cm}\caption*{\tiny person wear watch}
    \end{subfigure}\hfill
    \begin{subfigure}[t]{.23\linewidth}
        \centering
        \includegraphics[width=\linewidth]{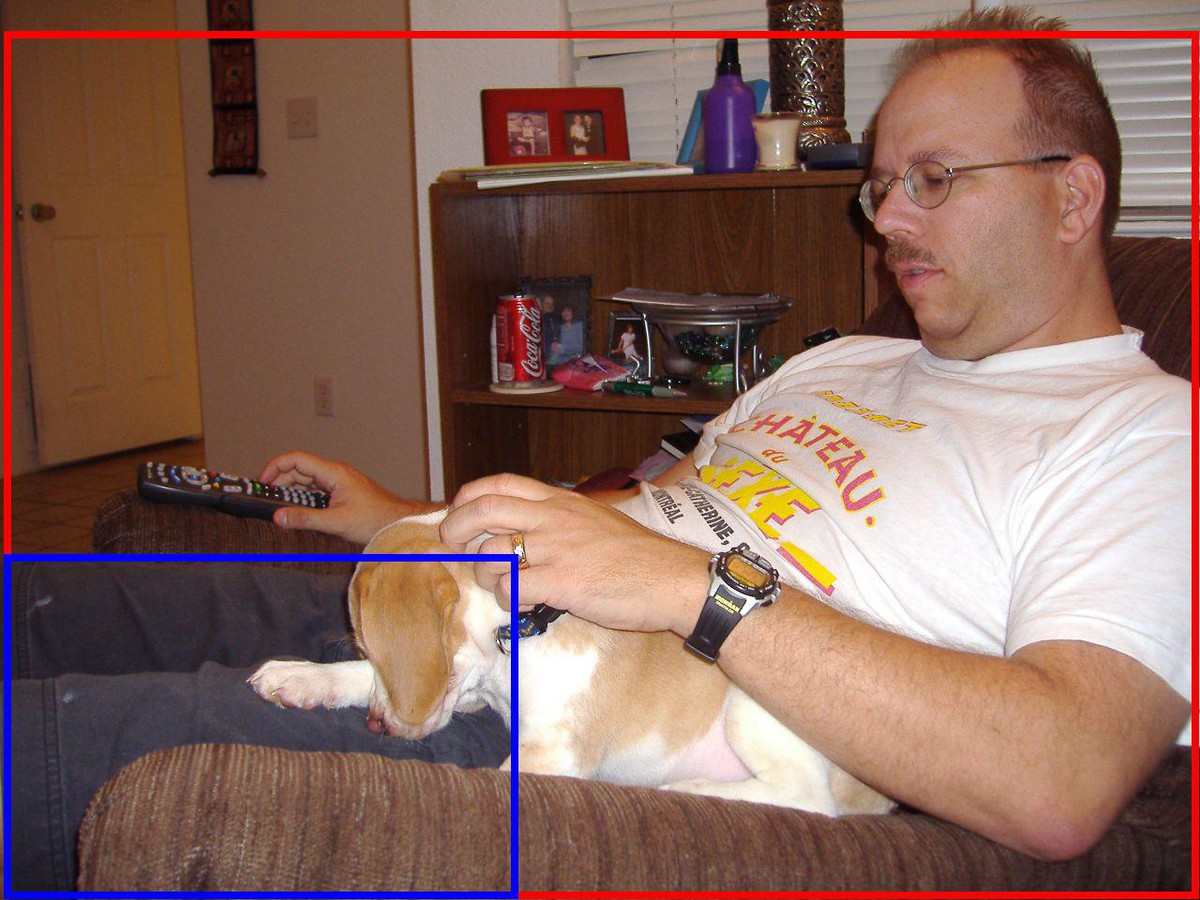}
        \vspace{-.6cm}\caption*{\tiny person wear pants}
    \end{subfigure}\hfill
    \begin{subfigure}[t]{.23\linewidth}
        \centering
        \includegraphics[width=\linewidth]{figures/hico/empty.png}
        \vspace{-.6cm}\caption*{\tiny (not possible with ground-truth objects)}
    \end{subfigure}\hfill
    \begin{subfigure}[t]{.23\linewidth}
        \centering
        \includegraphics[width=\linewidth]{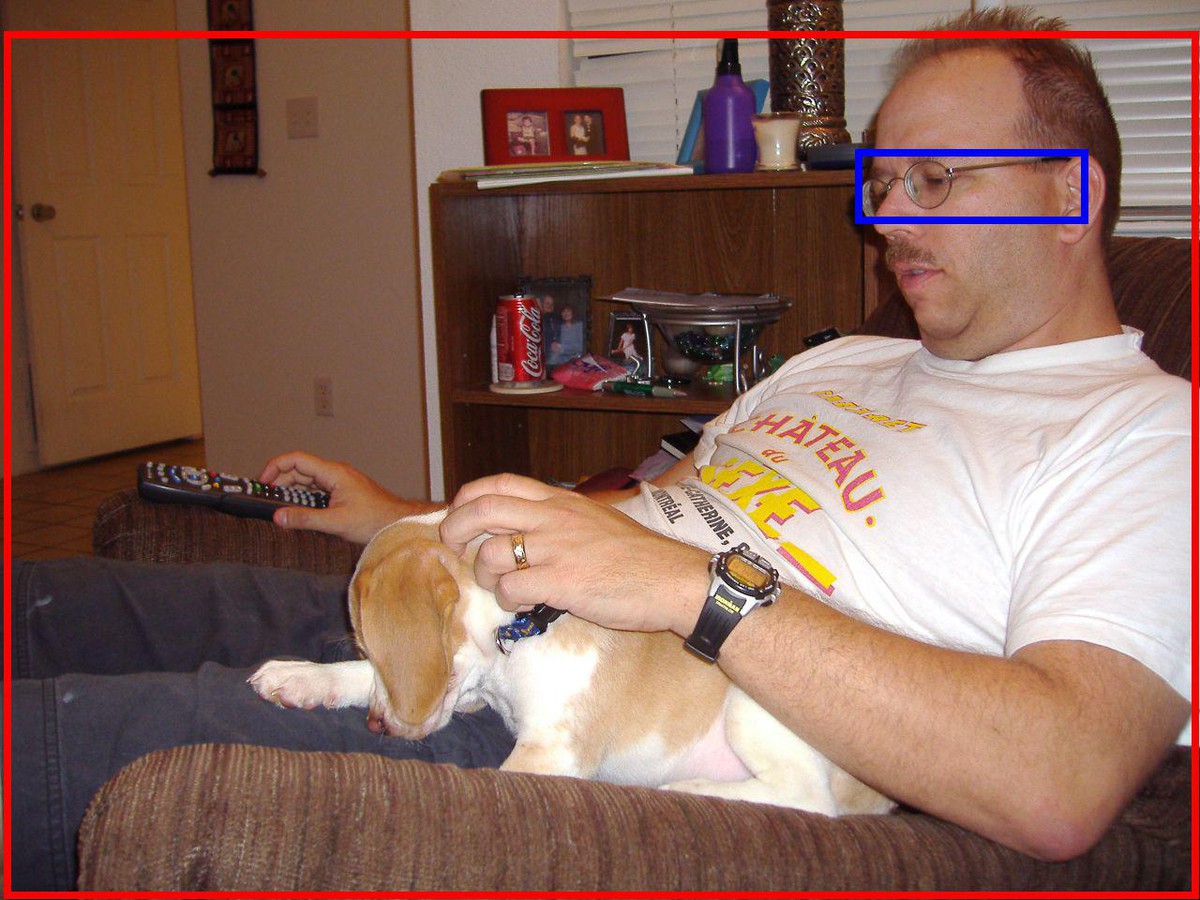}
        \vspace{-.6cm}\caption*{\tiny person hold glasses \\(wrong predicate)}
    \end{subfigure}\\
    \begin{subfigure}[t]{.23\linewidth}
        \centering
        \includegraphics[width=\linewidth]{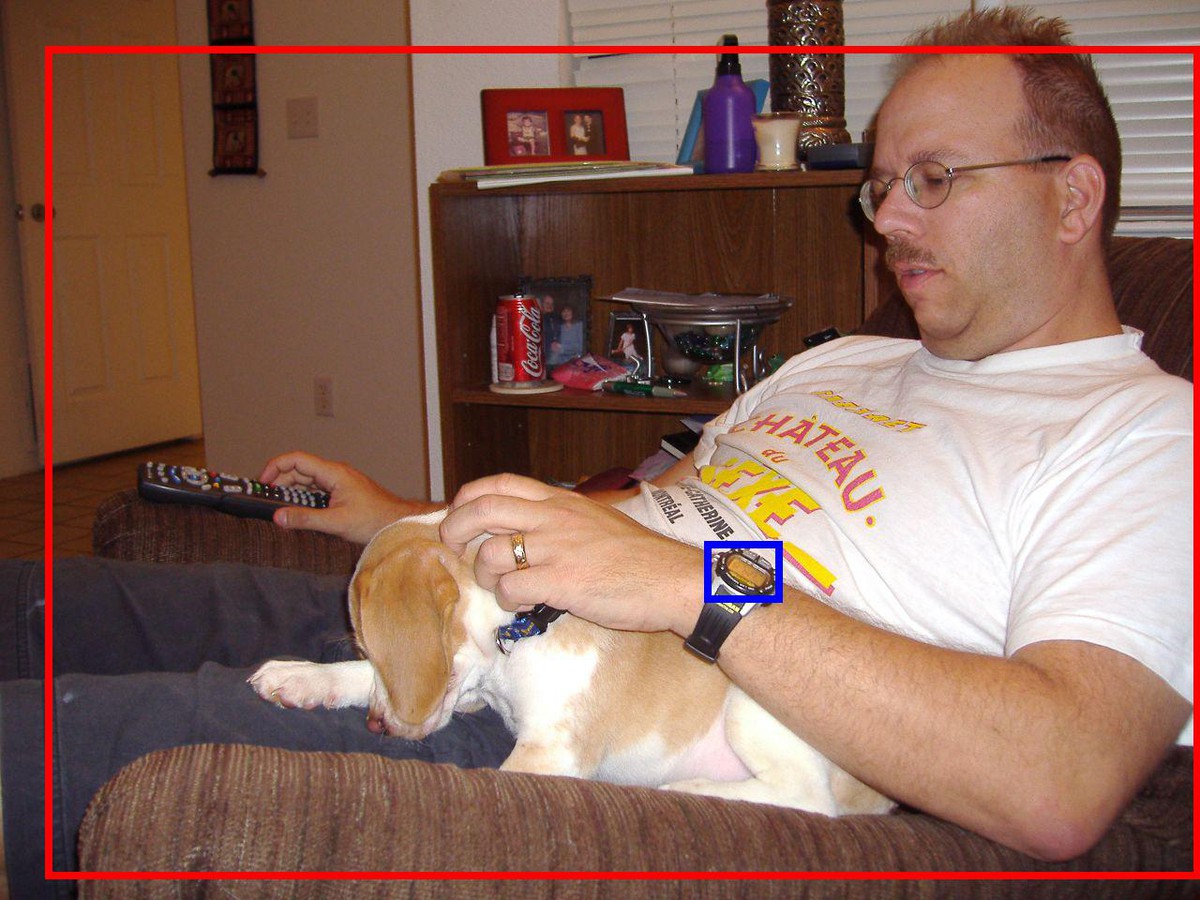}
        \vspace{-.6cm}\caption*{\tiny person wear watch}
    \end{subfigure}\hfill
    \begin{subfigure}[t]{.23\linewidth}
        \centering
        \includegraphics[width=\linewidth]{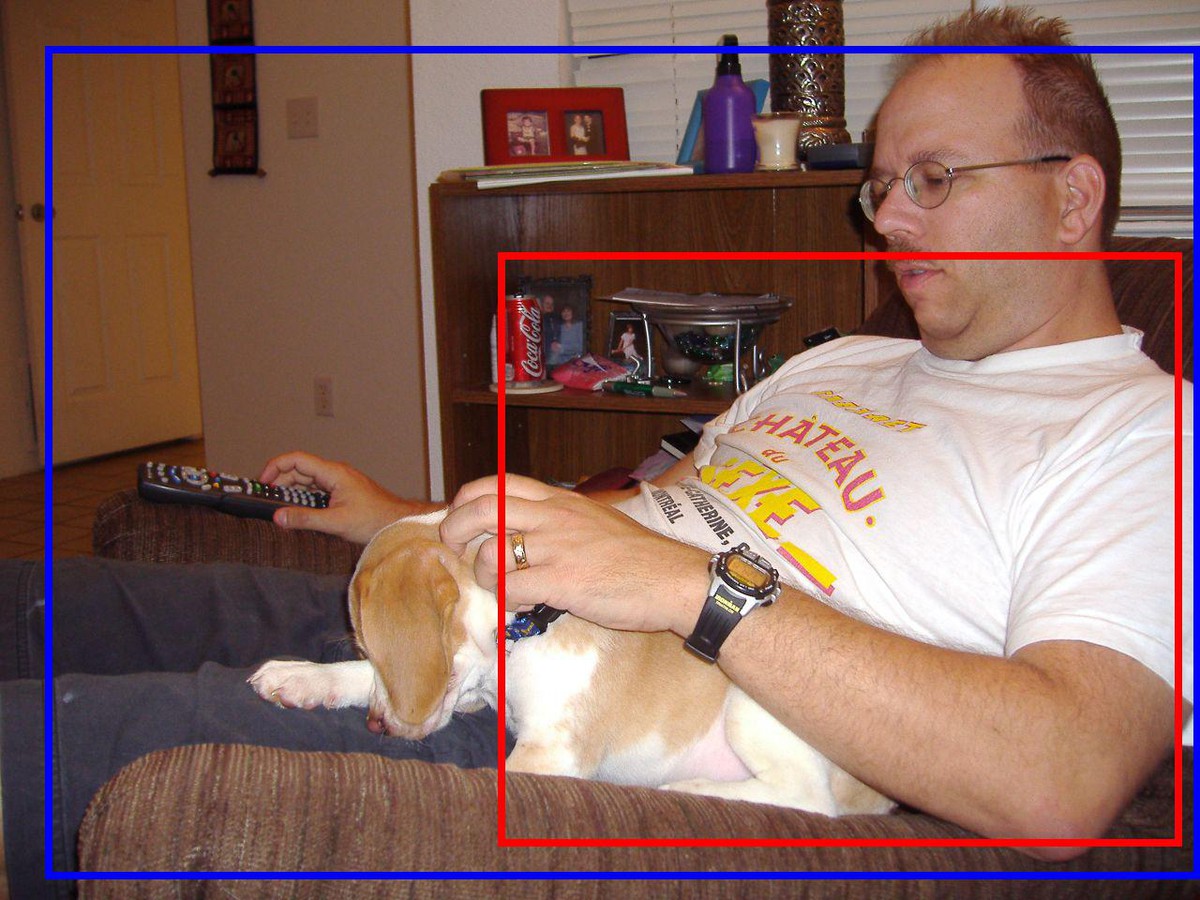}
        \vspace{-.6cm}\caption*{\tiny shirt on person}
    \end{subfigure}\hfill
    \begin{subfigure}[t]{.23\linewidth}
        \centering
        \includegraphics[width=\linewidth]{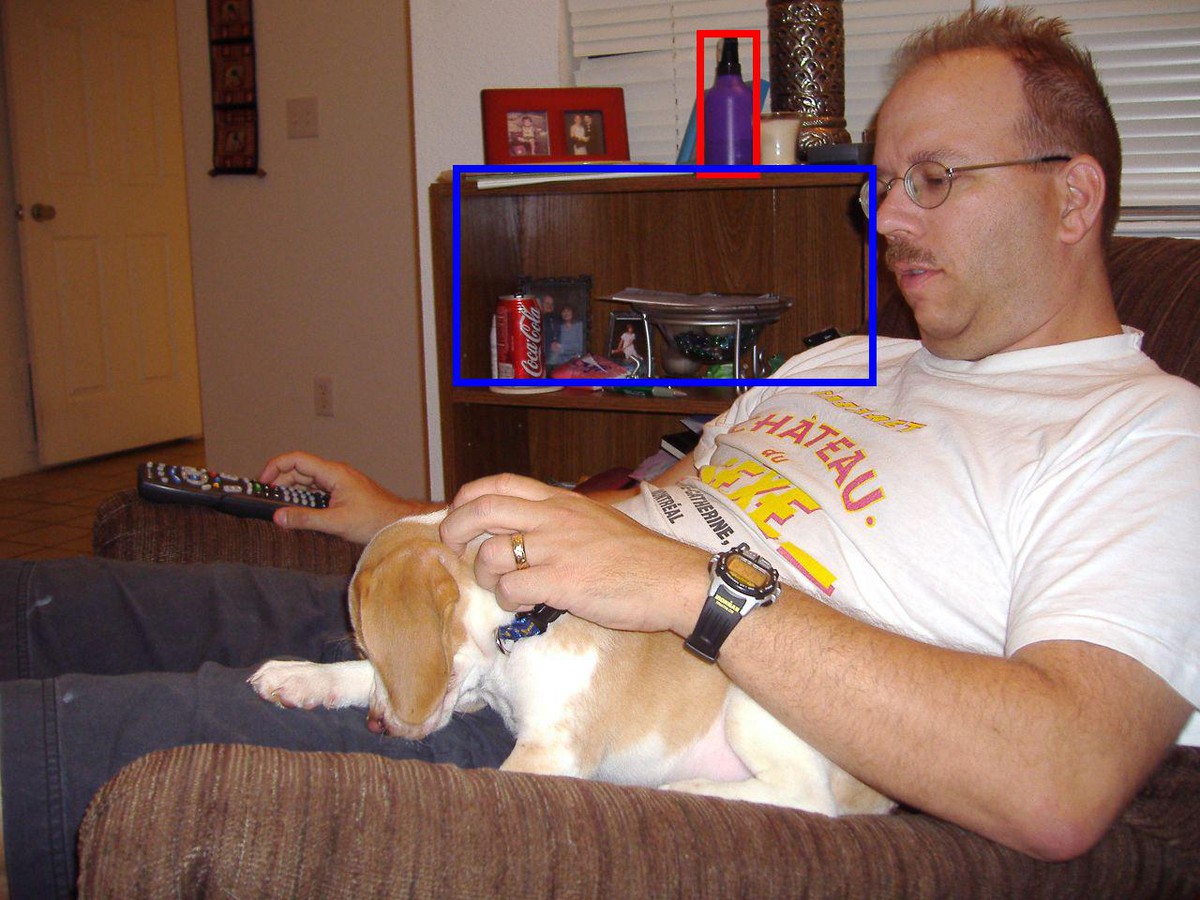}
        \vspace{-.6cm}\caption*{\tiny bottle on table\\(wrong object)}
    \end{subfigure}\hfill
    \begin{subfigure}[t]{.23\linewidth}
        \centering
        \includegraphics[width=\linewidth]{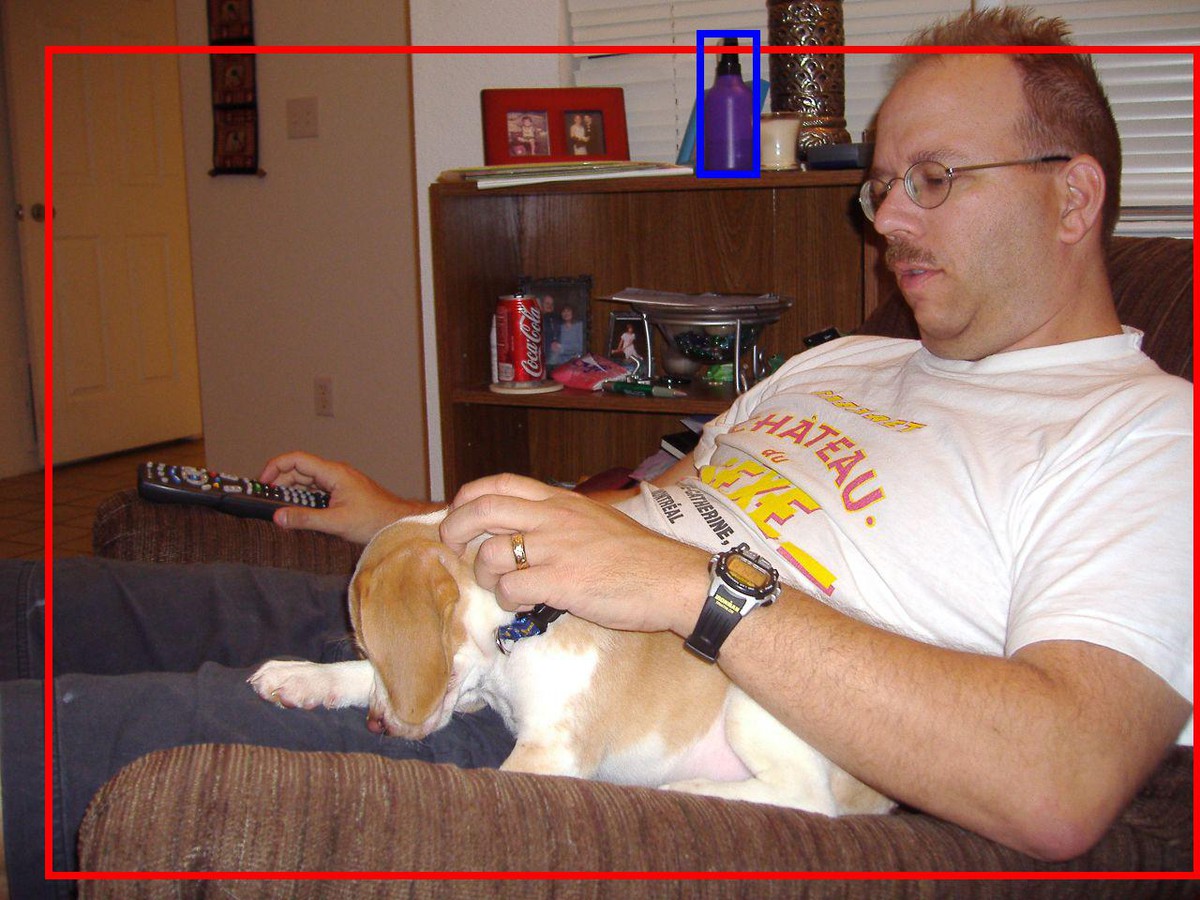}
        \vspace{-.6cm}\caption*{\tiny person hold bottle\\(no interaction)}
    \end{subfigure}\\
    \begin{subfigure}[t]{.23\linewidth}
        \centering
        \includegraphics[width=\linewidth]{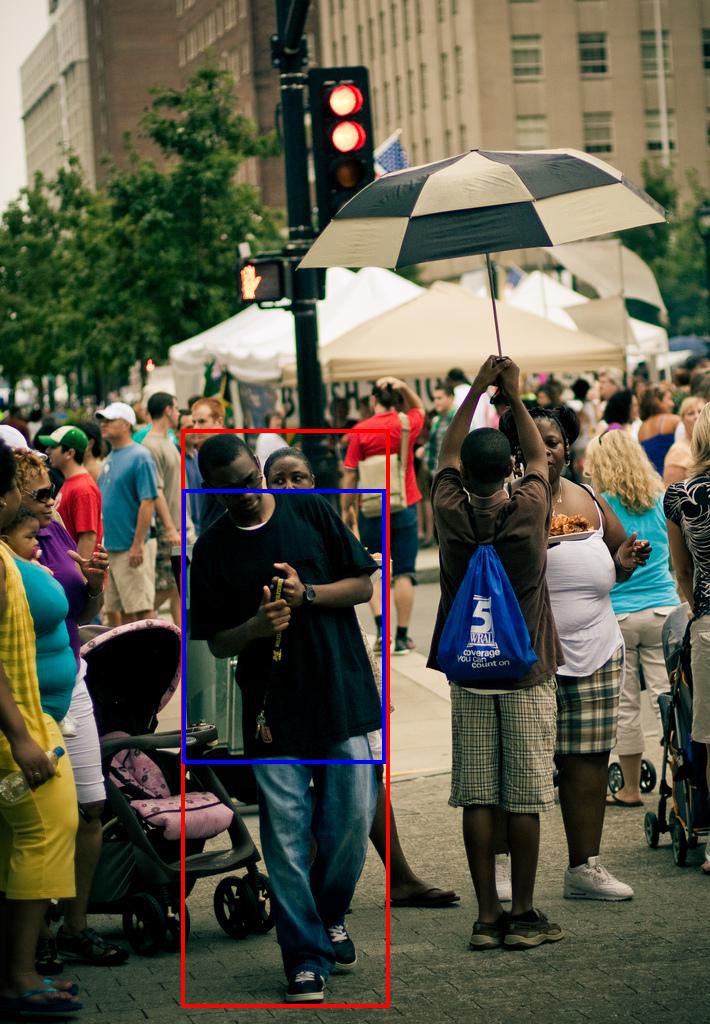}
        \vspace{-.6cm}\caption*{\tiny person wear shirt}
    \end{subfigure}\hfill
    \begin{subfigure}[t]{.23\linewidth}
        \centering
        \includegraphics[width=\linewidth]{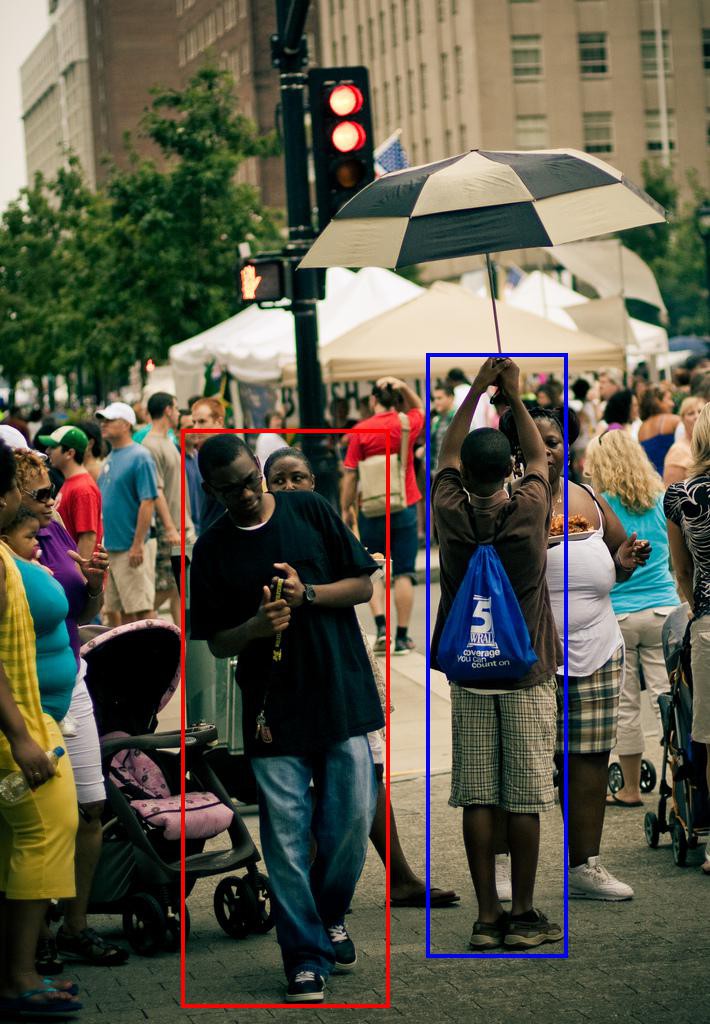}
        \vspace{-.6cm}\caption*{\tiny person near person}
    \end{subfigure}\hfill
    \begin{subfigure}[t]{.23\linewidth}
        \centering
        \includegraphics[width=\linewidth]{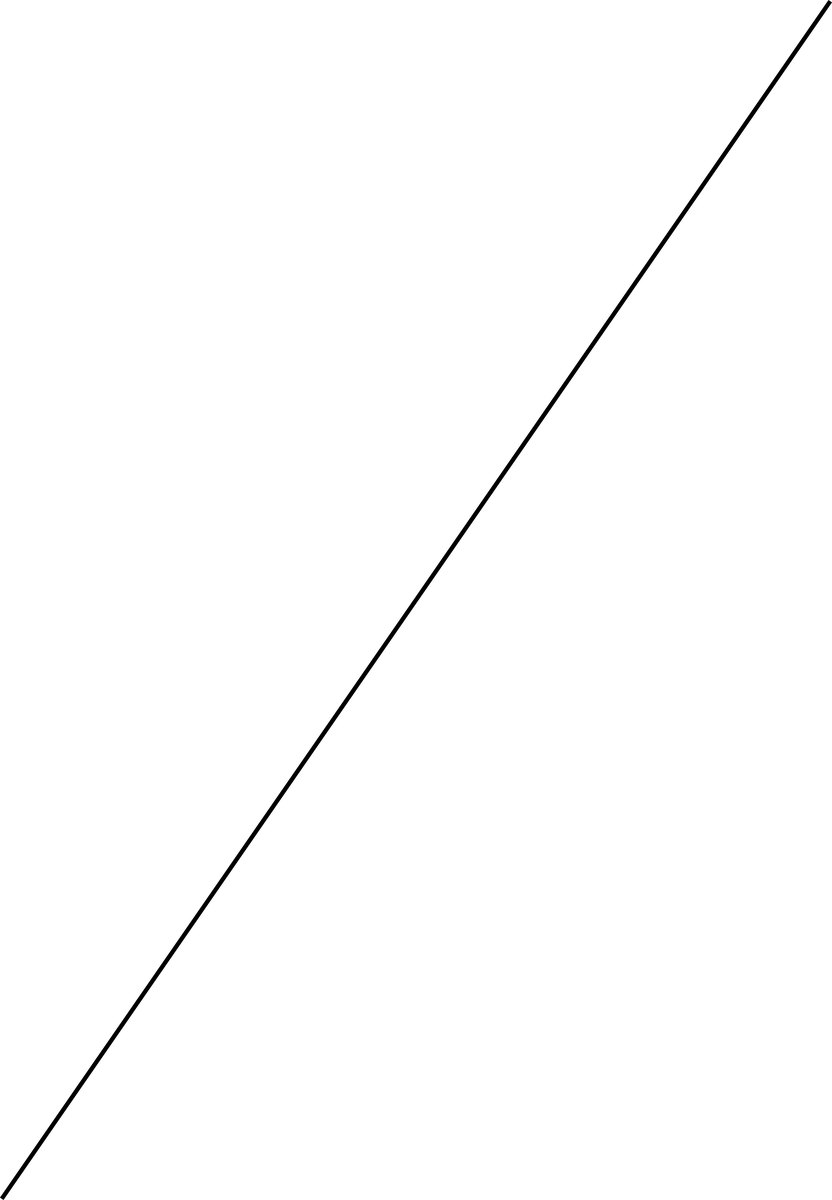}
        \vspace{-.6cm}\caption*{\tiny (not possible with ground-truth objects)}
    \end{subfigure}\hfill
    \begin{subfigure}[t]{.23\linewidth}
        \centering
        \includegraphics[width=\linewidth]{figures/additional/vrd/gt_boxes/4847012655_42e6263cfd_b_person-hold-umbrella_wrong_pair.jpg}
        \vspace{-.6cm}\caption*{\tiny person hold umbrella \\(wrong subj-obj pair)}
    \end{subfigure}\\
    \begin{subfigure}[t]{.23\linewidth}
        \centering
        \includegraphics[width=\linewidth]{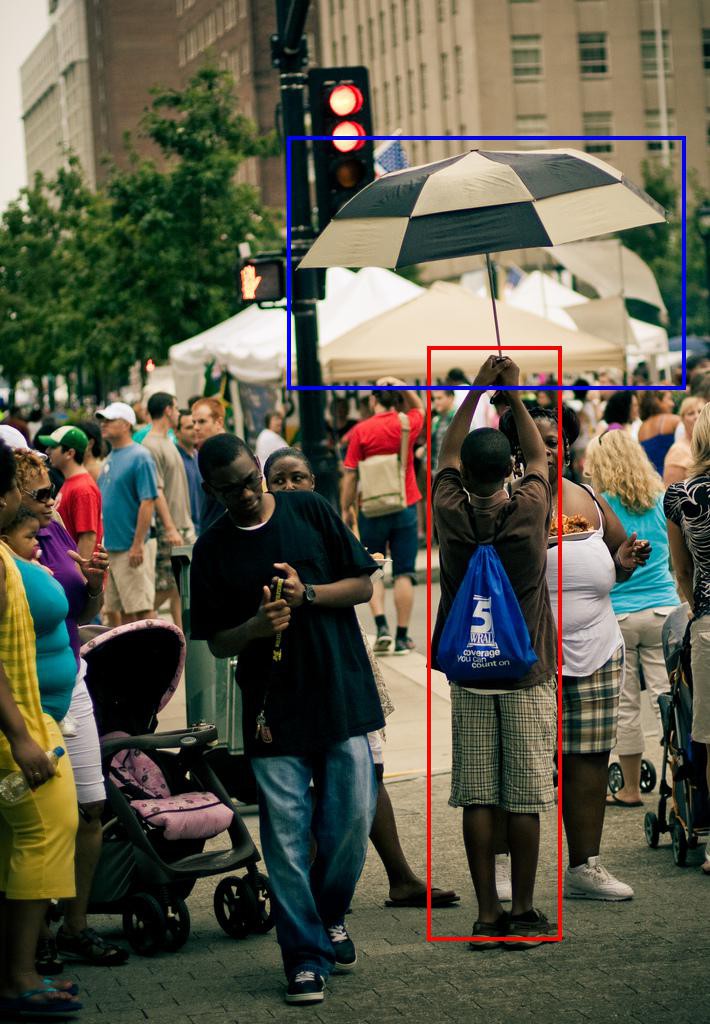}
        \vspace{-.6cm}\caption*{\tiny person under umbrella}
    \end{subfigure}\hfill
    \begin{subfigure}[t]{.23\linewidth}
        \centering
        \includegraphics[width=\linewidth]{figures/additional/vrd/d2_boxes/4847012655_42e6263cfd_b_person-on-street.jpg}
        \vspace{-.6cm}\caption*{\tiny person on street}
    \end{subfigure}\hfill
    \begin{subfigure}[t]{.23\linewidth}
        \centering
        \includegraphics[width=\linewidth]{figures/additional/vrd/empty.png}
        \vspace{-.6cm}\caption*{\tiny (none for this image)}
    \end{subfigure}\hfill
    \begin{subfigure}[t]{.23\linewidth}
        \centering
        \includegraphics[width=\linewidth]{figures/additional/vrd/d2_boxes/4847012655_42e6263cfd_b_person-wear-shirt_wrong_pair.jpg}
        \vspace{-.6cm}\caption*{\tiny person wear shirt \\(wrong subj-obj pair)}
    \end{subfigure}    
    \caption{\small \textbf{Additional detections on VRD.} Odd rows use ground-truth objects, even rows use Faster R-CNN objects. Subjects are framed in red, objects in blue. Left to right: correct relationship detection, correct but missing ground-truth, incorrect due to object misdetection, incorrect detection. Images are chosen at random from the test set, all depicted triplets are selected from the top 10 detections of an image}
    \label{fig:app-additional-results-vrd}
\end{figure}

\begin{figure}[ht]
    \centering
    \captionsetup[subfigure]{justification=centering}
    \begin{subfigure}[t]{.23\linewidth}
        \centering
        \includegraphics[width=\linewidth]{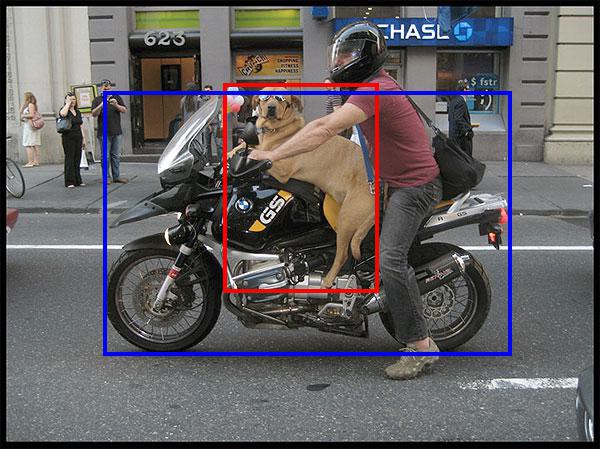}
        \vspace{-.6cm}\caption*{\tiny dog ride motorcycle}
    \end{subfigure}\hfill
    \begin{subfigure}[t]{.23\linewidth}
        \centering
        \includegraphics[width=\linewidth]{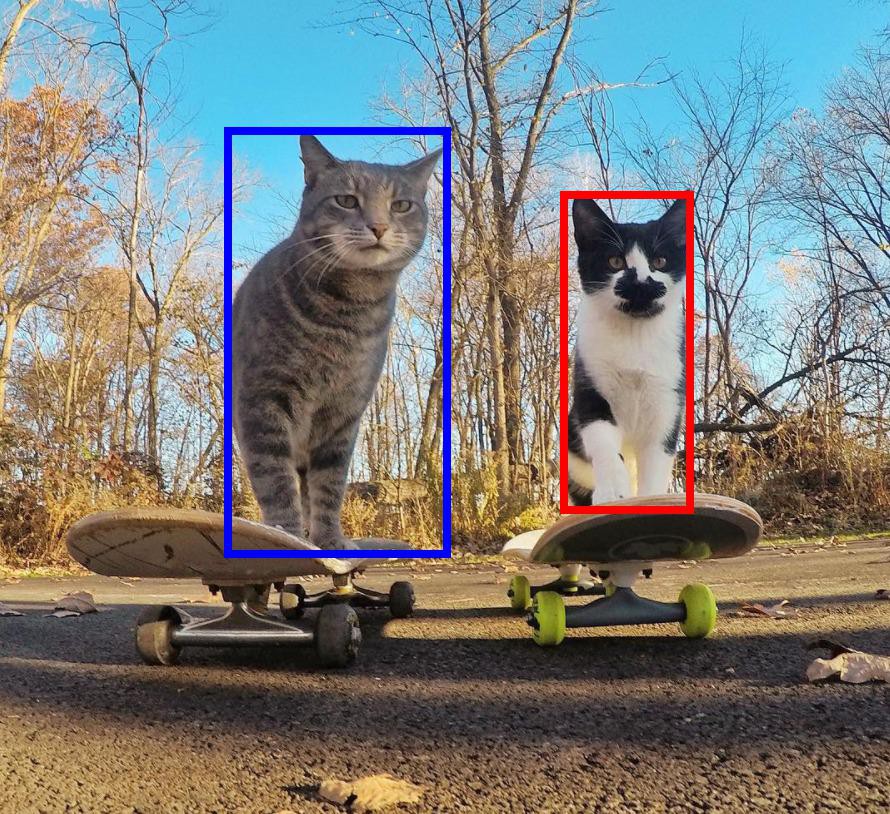}
        \vspace{-.6cm}\caption*{\tiny cat next to cat}
    \end{subfigure}\hfill
    \begin{subfigure}[t]{.23\linewidth}
        \centering
        \includegraphics[width=\linewidth]{figures/hico/empty.png}
        \vspace{-.6cm}\caption*{\tiny (not possible with ground-truth objects)}
    \end{subfigure}\hfill
    \begin{subfigure}[t]{.23\linewidth}
        \centering
        \includegraphics[width=\linewidth]{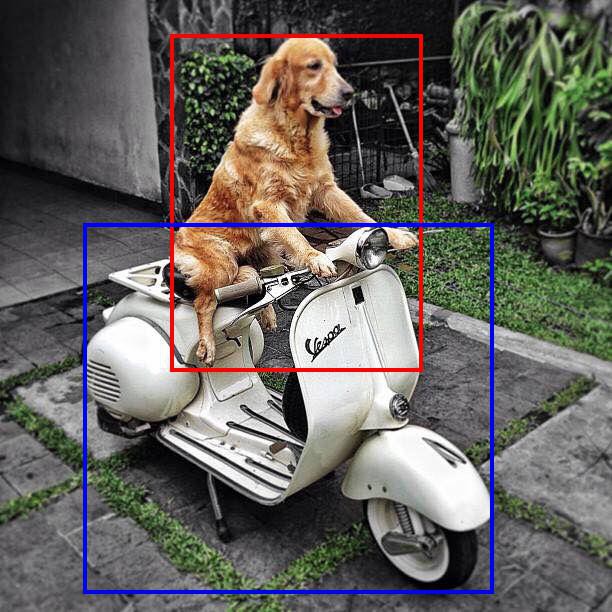}
        \vspace{-.6cm}\caption*{\tiny dog under motorcycle \\(wrong predicate)}
    \end{subfigure}\\
    \begin{subfigure}[t]{.23\linewidth}
        \centering
        \includegraphics[width=\linewidth]{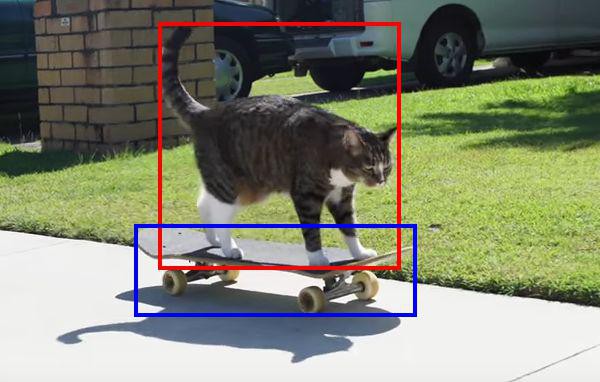}
        \vspace{-.6cm}\caption*{\tiny cat ride skateboard}
    \end{subfigure}\hfill
    \begin{subfigure}[t]{.23\linewidth}
        \centering
        \includegraphics[width=\linewidth]{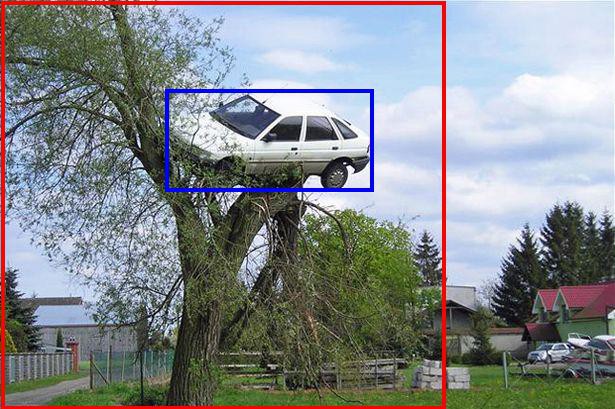}
        \vspace{-.6cm}\caption*{\tiny tree behind car}
    \end{subfigure}\hfill
    \begin{subfigure}[t]{.23\linewidth}
        \centering
        \includegraphics[width=\linewidth]{figures/hico/empty.png}
        \vspace{-.6cm}\caption*{\tiny (not possible with ground-truth objects)}
    \end{subfigure}\hfill
    \begin{subfigure}[t]{.23\linewidth}
        \centering
        \includegraphics[width=\linewidth]{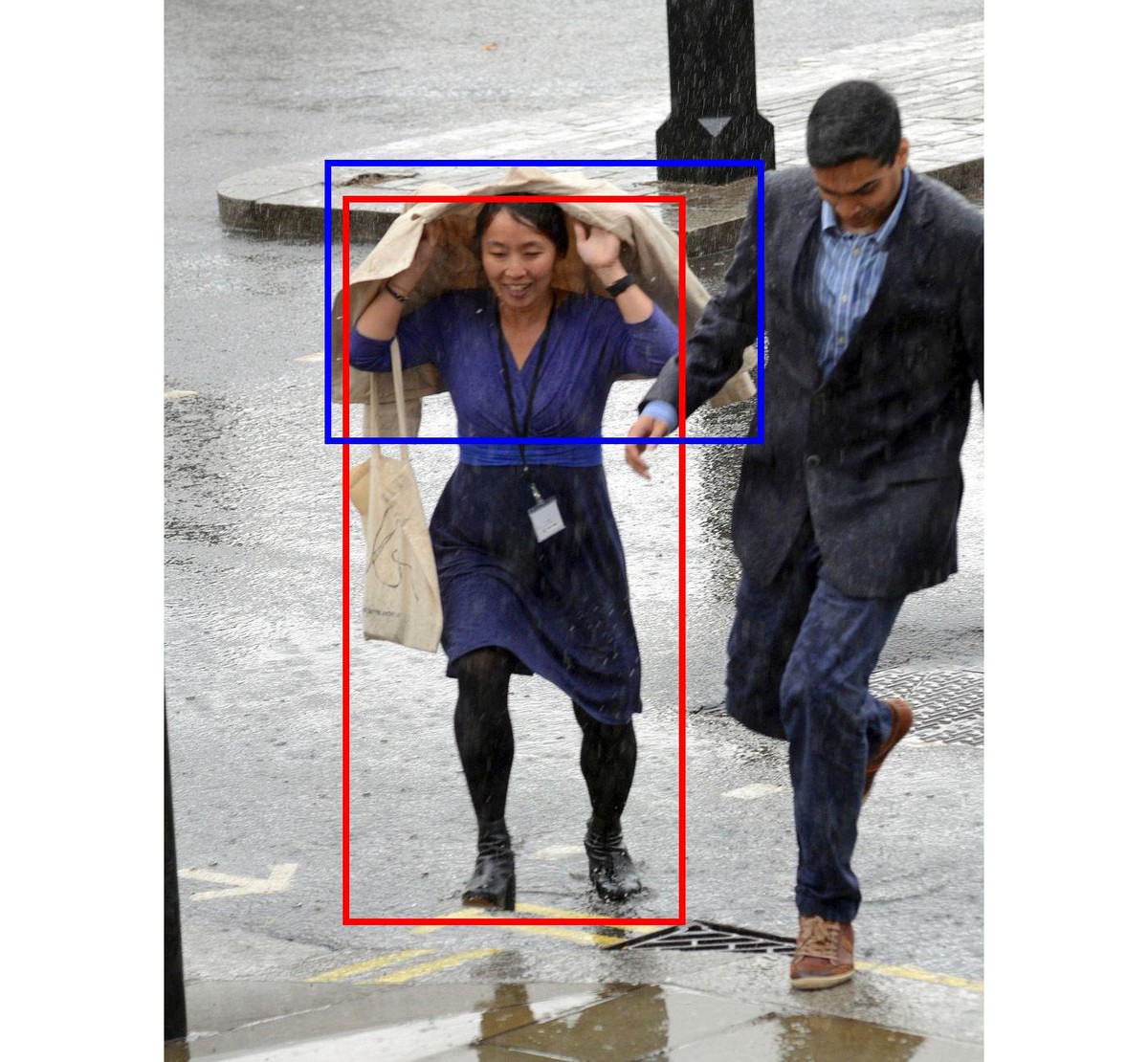}
        \vspace{-.6cm}\caption*{\tiny person wear jacket\\(wrong predicate)}
    \end{subfigure}\\
    \begin{subfigure}[t]{.23\linewidth}
        \centering
        \includegraphics[width=\linewidth]{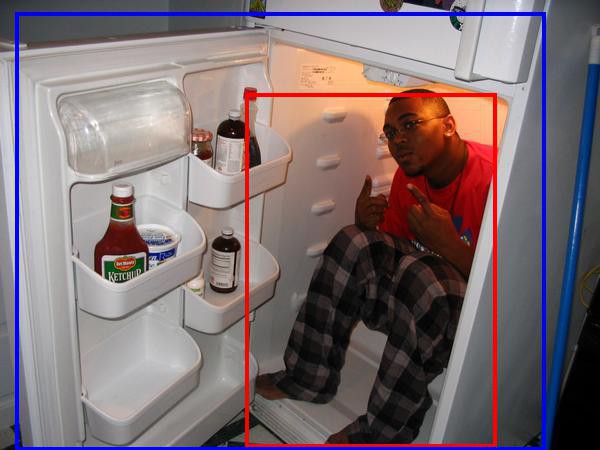}
        \vspace{-.6cm}\caption*{\tiny person in refrigerator}
    \end{subfigure}\hfill
    \begin{subfigure}[t]{.23\linewidth}
        \centering
        \includegraphics[width=\linewidth]{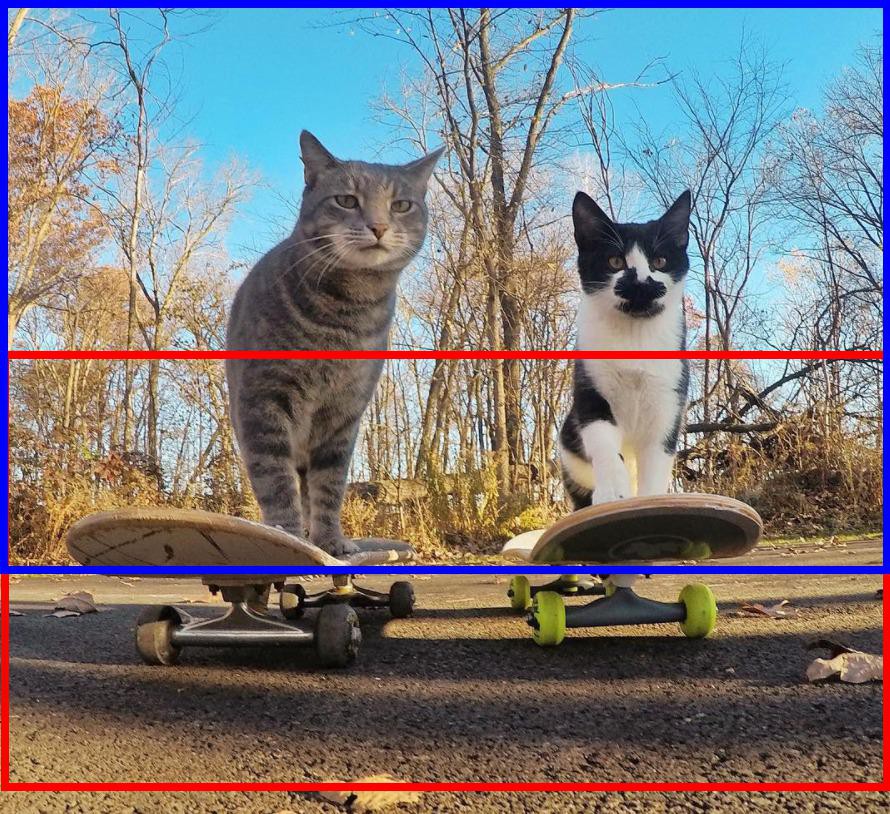}
        \vspace{-.6cm}\caption*{\tiny street under trees}
    \end{subfigure}\hfill
    \begin{subfigure}[t]{.23\linewidth}
        \centering
        \includegraphics[width=\linewidth]{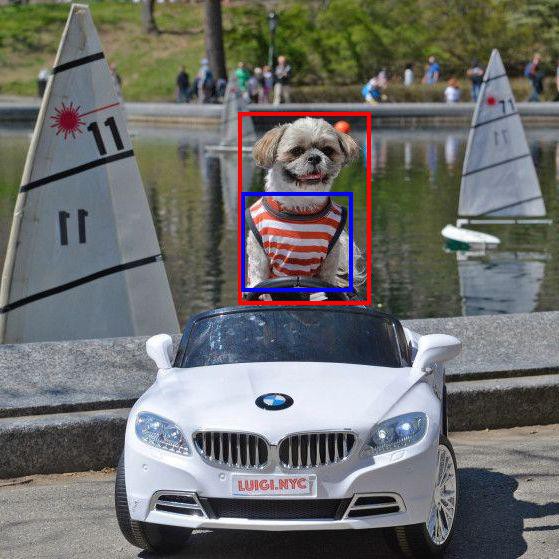}
        \vspace{-.6cm}\caption*{\tiny person wear shirt (correct subj: dog)}
    \end{subfigure}\hfill
    \begin{subfigure}[t]{.23\linewidth}
        \centering
        \includegraphics[width=\linewidth]{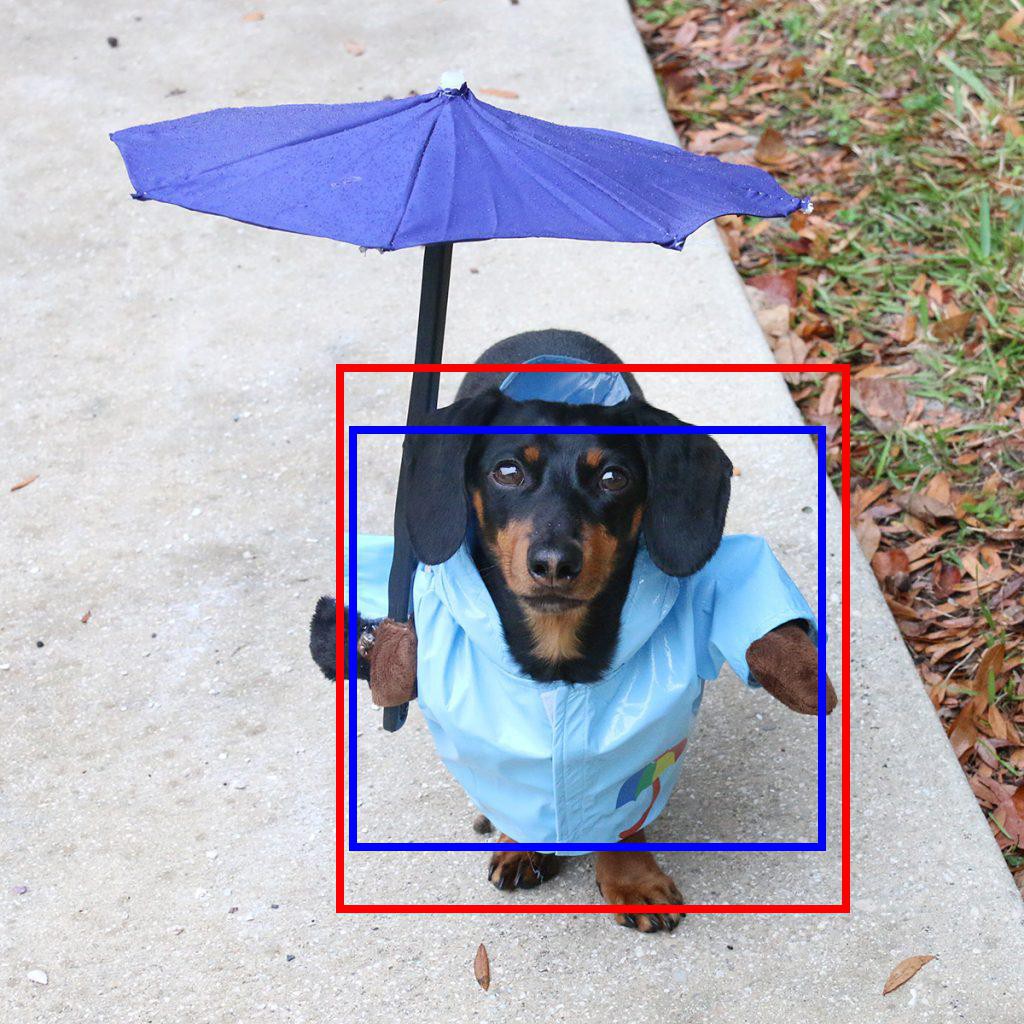}
        \vspace{-.6cm}\caption*{\tiny dog next to jacket\\(wrong predicate)}
    \end{subfigure}\\
    \begin{subfigure}[t]{.23\linewidth}
        \centering
        \includegraphics[width=\linewidth]{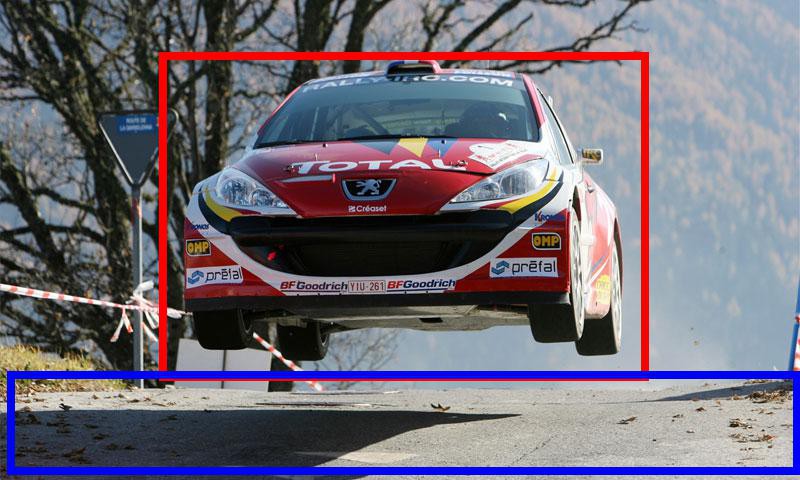}
        \vspace{-.6cm}\caption*{\tiny car above street}
    \end{subfigure}\hfill
    \begin{subfigure}[t]{.23\linewidth}
        \centering
        \includegraphics[width=\linewidth]{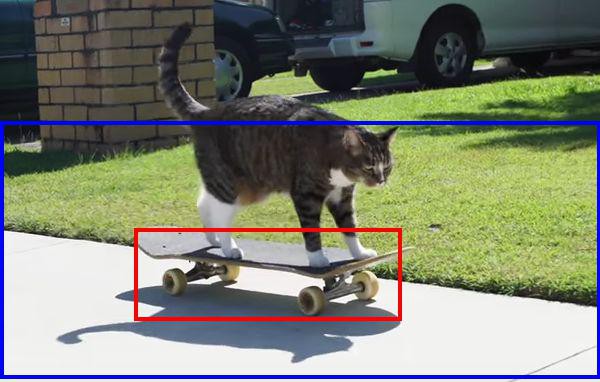}
        \vspace{-.6cm}\caption*{\tiny skateboard above street}
    \end{subfigure}\hfill
    \begin{subfigure}[t]{.23\linewidth}
        \centering
        \includegraphics[width=\linewidth]{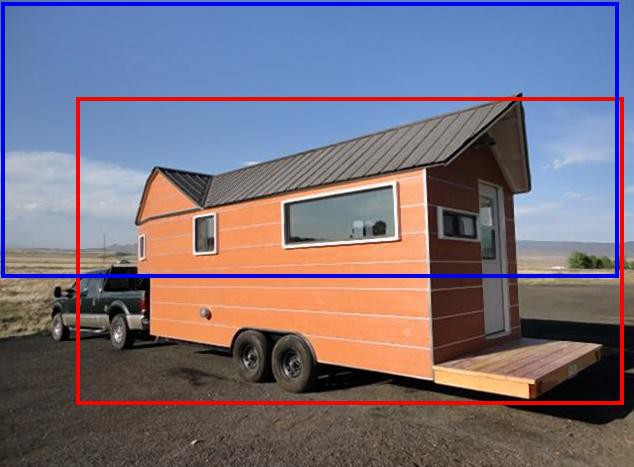}
        \vspace{-.6cm}\caption*{\tiny truck below sky}
    \end{subfigure}\hfill
    \begin{subfigure}[t]{.23\linewidth}
        \centering
        \includegraphics[width=\linewidth]{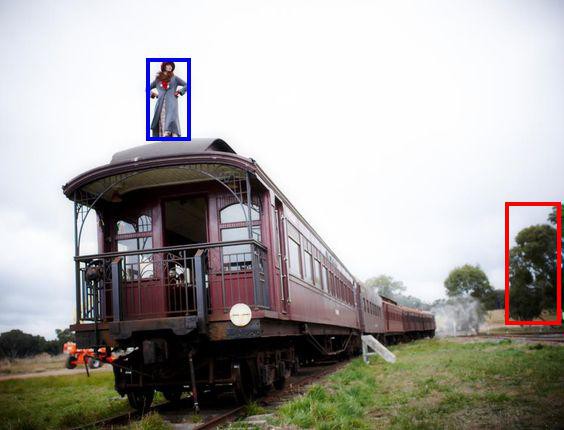}
        \vspace{-.6cm}\caption*{\tiny tree next to person \\(wrong predicate)}
    \end{subfigure}    
    \caption{\small \textbf{Additional detections on UnRel.} Top two rows use ground-truth objects, bottom two rows use Faster R-CNN objects. Subjects are framed in red, objects in blue. Left to right: correct relationship detection, correct but missing ground-truth, incorrect due to object misdetection, incorrect detection. Images are chosen at random from the test set, all depicted triplets are selected from the top 25 detections}
    \label{fig:app-additional-results-unrel}
\end{figure}
\fi

\end{document}